\documentclass{article}

\usepackage{arxiv}

\usepackage[utf8]{inputenc} 
\usepackage[T1]{fontenc}    
\usepackage{hyperref}       
\usepackage{url}            
\usepackage{booktabs}       
\usepackage{amsfonts}       
\usepackage{nicefrac}       
\usepackage{microtype}      
\usepackage{lipsum}
\usepackage{graphicx}

\usepackage[square,numbers]{natbib} 
\usepackage{amsfonts}
\usepackage{soul}
\usepackage{textcomp}
\usepackage{setspace}
\usepackage{placeins}
\usepackage{tabularx}
\usepackage{multirow}
\usepackage{adjustbox}
\usepackage{tabulary}
\usepackage{textcomp}
\usepackage{mathtools}
\usepackage{algorithm}
\usepackage{subcaption}
\usepackage{amsmath}
\usepackage{mwe}
\usepackage{bm}
\usepackage[noend]{algpseudocode}
\usepackage{parskip}
\usepackage{mwe}
\title{Geometric Change Detection in Digital Twins using 3D Machine Learning}

\author{
 Tiril Sundby \\
Department of Engineering Cybernetics\\
Norwegian University of Science and Technology\\
Trondheim, Norway \\
  \texttt{tirils@live.no} \\
   \And
Julia Maria Graham \\
Department of Engineering Cybernetics\\
Norwegian University of Science and Technology\\
Trondheim, Norway \\
  \texttt{julia.graham@live.no} \\
  \And
Adil Rasheed \\
Department of Engineering Cybernetics\\
Norwegian University of Science and Technology\\
Trondheim, Norway \\
  \texttt{adil.rasheed@ntnu.no} \\
  
    \And
Mandar Tabib \\
Mathematics and Cybernetics\\
Norwegian University of Science and Technology\\
Trondheim, Norway \\
  \texttt{mandar.tabib@sintef.no} \\
      \And
Omer San \\
School of Mechanical and Aerospace Engineering\\
Oklahoma State University\\
Oklahoma 74078 USA\\
\texttt{osan@okstate.edu} \\
}

\begin{document}
\maketitle
\begin{abstract}
Digital twins are meant to bridge the gap between real-world physical systems and virtual representations. Both stand-alone and descriptive digital twins incorporate 3D geometric models, which are the physical representations of objects in the digital replica. Digital twin applications are required to rapidly update internal parameters with the evolution of their physical counterpart. Due to an essential need for having high-quality geometric models for accurate physical representations, the storage and bandwidth requirements for storing 3D model information can quickly exceed the available storage and bandwidth capacity. In this work, we demonstrate a novel approach to geometric change detection in the context of a digital twin. We address the issue through a combined solution of Dynamic Mode Decomposition (DMD) for motion detection, YOLOv5 for object detection, and 3D machine learning for pose estimation. DMD is applied for background subtraction, enabling detection of moving foreground objects in real-time. The video frames containing detected motion are extracted and used as input to the change detection network. The object detection algorithm YOLOv5 is applied to extract the bounding boxes of detected objects in the video frames. Furthermore, the rotational pose of each object is estimated in a 3D pose estimation network. A series of convolutional neural networks (CNNs) conducts feature extraction from images and 3D model shapes. Then, the network outputs the estimated Euler angles of the camera orientation with respect to the object in the input image. By only storing data associated with a detected change in pose, we minimize necessary storage and bandwidth requirements while still being able to recreate the 3D scene on demand. Our assessment of the new geometric detection framework shows that the proposed methodology could represent a viable tool in emerging digital twin applications.
\end{abstract}


\section{Introduction}
With the recent age of digitalization the world is quickly embracing new technologies like digital twin which is seen as an enabler for many of the autonomous systems. A digital twin \cite{rasheed_digital_2020} is defined as a virtual representation of a physical asset enabled through data and simulators for real-time prediction, optimization, monitoring, controlling, and improved decision making. The capability of DT can be ranked on a scale from 0-5 (0-standalone, 1-descriptive, 2-diagnostic, 3-predictive, 4-prescriptive, 5-autonomy) as shown in \autoref{fig:DTcapability}. A standalone or descriptive digital twin consists of computer-aided design (CAD) models which are the physical representations of the objects building it \citep{zheng_application_2019}. CAD models based on the accuracy requirements in a digital twin can be enormous in size. For a digital twin to achieve a capability level of 5 ie. full autonomy it will require the ability to understand the 3D environment in real time and to keep track of the changes. The requirements of frequently updating the digital twin with the evolution of the physical assets leads to a situation wherein - the storage and bandwidth requirements for archiving the CAD models as a function of time will quickly exceed the available storage \citep{philip_chen_data-intensive_2014} and communication bandwidth capacity \citep{minerva_digital_2020, west_is_2017}. Fortunately major advancements have been made in the field of motion detection, object detection and classification, and pose estimation which can address these challenges. A brief description of the methodologies in the context of our workflow are described below: 
\begin{figure*}
    \includegraphics[width=\linewidth]{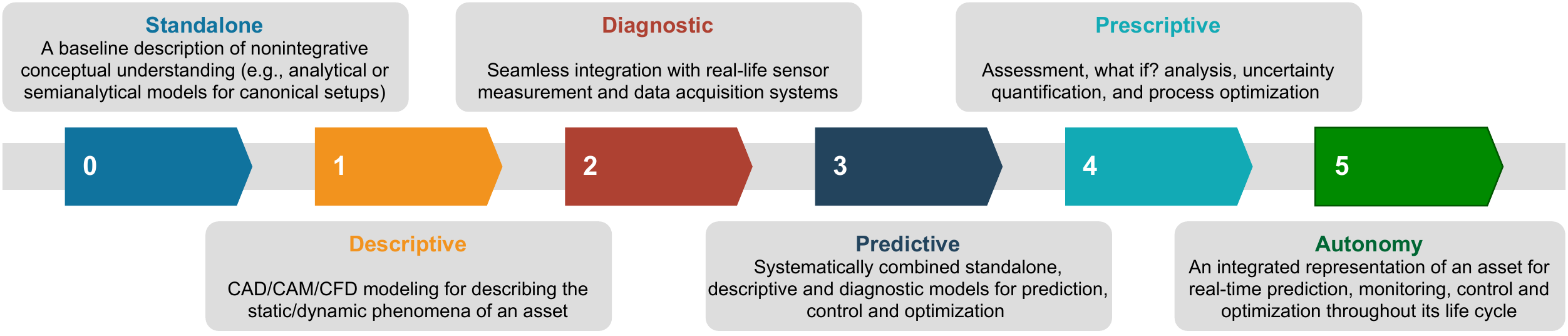}
    \caption{Digital twin capabilities, on a scale from 0-5}
    \label{fig:DTcapability}
\end{figure*}

\textit{Motion detection:} Traditional motion detection approaches can be categorized into forms such as background subtraction, frame differencing, temporal differencing and optical flow \cite{7087138}.  Amongst these, the background subtraction method is the most reliable method that involves initializing a background model first, and then a comparison between each pixel of the current frame with the assumed background model color map provides information on whether the pixel belongs to the background or the foreground. If the difference between colors is less than a chosen threshold, then the pixel is considered to be belonging to the background, else it belongs to the foreground. Typical background detection method comprises of  eigen backgrounds, mixture of Gaussians, concurrence of image variations,  Running Gaussian average, Kernel Density Estimation (KDE), Sequential KD approximation and temporal median filter (overviews of many of these approaches are provided by Bouwmans \cite{BOUWMANS201431} and Sobral and Vacavant\cite{SOBRAL20144}). Many of these methods face  challenges with camera jitter, illumination changes, shadows and dynamic backgrounds, and there is no single method currently available that is capable of handling all the challenges in real time without suffering performance failures. From the perspective of digital twin set-up, real-time efficiency of motion detection with adequate accuracy matters a lot.  In this regards, the authors in \cite{ISI:000380434700112,ISI:000425239601104,ISI:000489318600010} proposed the use of dynamic mode decomposition (DMD) and its variants: compressing dynamic mode decomposition (CDMD), multi-resoluton DMD (MRDMD) as novel ways of background-subtraction for motion detection approach. They demonstrated that the DMD worked robustly in real time using just a personal laptop-class computing power with adequate accuracy. Hence, for our proposed methodology of geometric change detection, we employ DMD as the preferred algorithm in our workflow for motion detection. 

\textit{Object detection and classification:} Conventional object detection involves separate stages comprising of: (a) object localization to find where the object lies in a frame which involves steps like informative region selection and feature extraction, and (b) followed by a final object classification stage. Most of the traditional approaches have been computationally inefficient as they use multi-scale sliding window in region selection and manual non-automated methods for feature extraction to find the object location in the image. With the advent of deep learning, the object detection process became more accurate and more efficient as  entire input image is fed to convolutional neural networks (CNN) to generate relevant feature map automatically. The feature maps were then used to identify regions and object classification \cite{ODISI:000494702100001,ODISI:000364047000005}. Although possible this used to be a computationally expensive process. However, with the development of You Only Look Once (YOLO) models \cite{OD7780460} which enable prediction of bounding boxes and class probabilities directly from entire images in just one evaluation, such tasks are now possible in real time. For this reason YOLO model is chosen as our preferred approach to object detection and classification in our workflow. 

\textit{Pose estimation:} Traditional methods to estimate the pose of a given 3D shape in an image involves establishing the correspondences between an object image and an object model, and for known object-shapes, these methods can  generally be categorized into feature-matching (involving key-point detection) methods and template-matching methods. Feature-matching approach refers to method of extracting local features or key-points from the image, matching them to the given 3D object model and then using a perspective-n-point (PnP) algorithm to obtain the 6D (3D translation and 3D rotations) pose based on estimated 2D-to-3D correspondences. These methods perform well on textured objects, but usually struggle with poorly-textured objects and low-resolution images. To deal with these types of objects, template-matching methods try to match the observed object to a stored template. But these methods struggle in cluttered environments subjected to partial occlusion or object truncation. More information on such method can be found in reviews \cite{PEsahin2020review,PEHe_2020}. In recent times, CNN based methods have shown robustness to the challenges faced by the traditional methods, and they have been developed to enable pose estimation from single image. These methods either treat the pose-estimation problem as classification-based or regression-based or as combination of the two and work by either (a) detecting pose of an object from image by 3D localization and rotation estimation \cite{PE7410693,PExiang2018posecnn},  or (b) by detecting key-point locations in images \cite{PEKPoberweger2018making,PEKPpavlakos20176dof,PEKPpeng2018pvnet}  (which helped in pose estimation through PnP) . However, most deep pose estimation methods are trained for specific object instances or categories (category-specific models). In the context of digital twin, it is desirable that we select a methodology that is robust to the environment and has the ability to generalize to newer unseen object categories and their orientations. In this direction, a recent category-free deep learning model (\citet{xiao_pose_2019}) estimates the pose of any object based on inputs of its image and its 3D shape (an instance-specific approach), and this model performs the pose-estimation well even if the object is dissimilar from the objects from its training phase. Hence, we employ this pose estimation in our proposed workflow.  

Overall, based on the available state-of-the-art and needs of the digital twins, the current work proposes a workflow for geometric change detection. The workflow, consisting of motion detection (using DMD), object detection and classification (using YOLO), and pose estimation (using deep learning), will ensure that instead of tracking and storing the full 3D CAD model evolution over time we only need to archive bare minimum information (like the changes in pose) whenever a motion is detected. This information can be used to apply transformations on the original state of the CAD models to recreate the state of these models at any point in time. In order to demonstrate the applicability of the workflow we developed a dedicated experimental setup to test it.  

The \autoref{sec:materialsandmethods} gives a detailed description of the proposed workflow, experimental setup, algorithms, software and hardware framework, and the data generation process. The results and discussions are then presented in \autoref{sec:resultsanddiscussion}. Finally, the main conclusions and potential extension of the work is presented in \autoref{sec:conclusionandfuturework}.

\section{Materials and Methods}
\label{sec:materialsandmethods}
\begin{figure}
    \makebox[\linewidth][c]{\includegraphics[width=\linewidth]{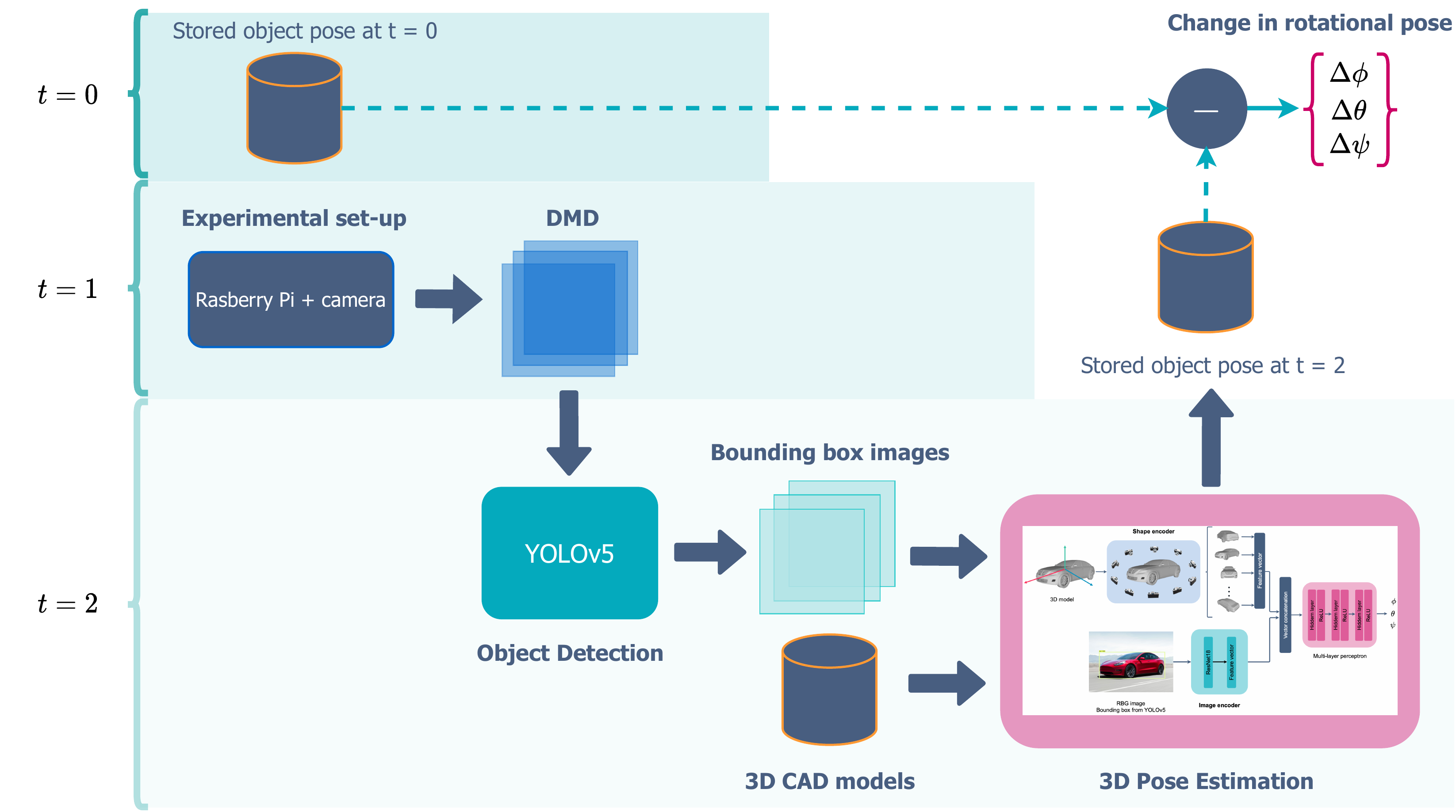}}
    \caption{Workflow for geometric change detection}
    \label{fig:full_architecture}
\end{figure}
The proposed workflow for geometric changes detection is presented in \autoref{fig:full_architecture}. We consider a cubical room consisting of 3D objects, each capable of moving with six degrees of freedom. 3D CAD models of these objects are saved at time $t=0$. An RGB camera mounted on Rasp Berry Pi4 is pointed towards the collection of objects. When the scene is stationary, the camera does not record anything. However, as soon as the objects start to move, the motion is detected using DMD. The whole sequence of the motion (between $t_{1}$ and $t_{2}$) is then recorded. When things become stationary again, the whole sequence is deleted after saving the last video frame containing the final change. The last frame is then analyzed using YOLO to detect objects and extract the corresponding bounding boxes which are then used as inputs to the pose estimation algorithms to estimate the effects of rotation ($\Delta \phi, \Delta \theta, \Delta \psi$). It should be pointed out that our current set-up is not suited for estimating the translation ($\Delta x, \Delta y, \Delta z$) as it would require some more modifications to the setup and algorithms. These are proposed as future work in the \autoref{sec:conclusionandfuturework}. In the following section, we give a brief overview of the implementation of the experimental setup, dataset generated using it, methods of motion detection, object detection and pose estimation employed in this work as well as the software, and hardware frameworks utilized. 

    \subsection{Experimental setup}
    \label{subsec:method_experimental_setup}
    \begin{figure}
        \includegraphics[width=\linewidth]{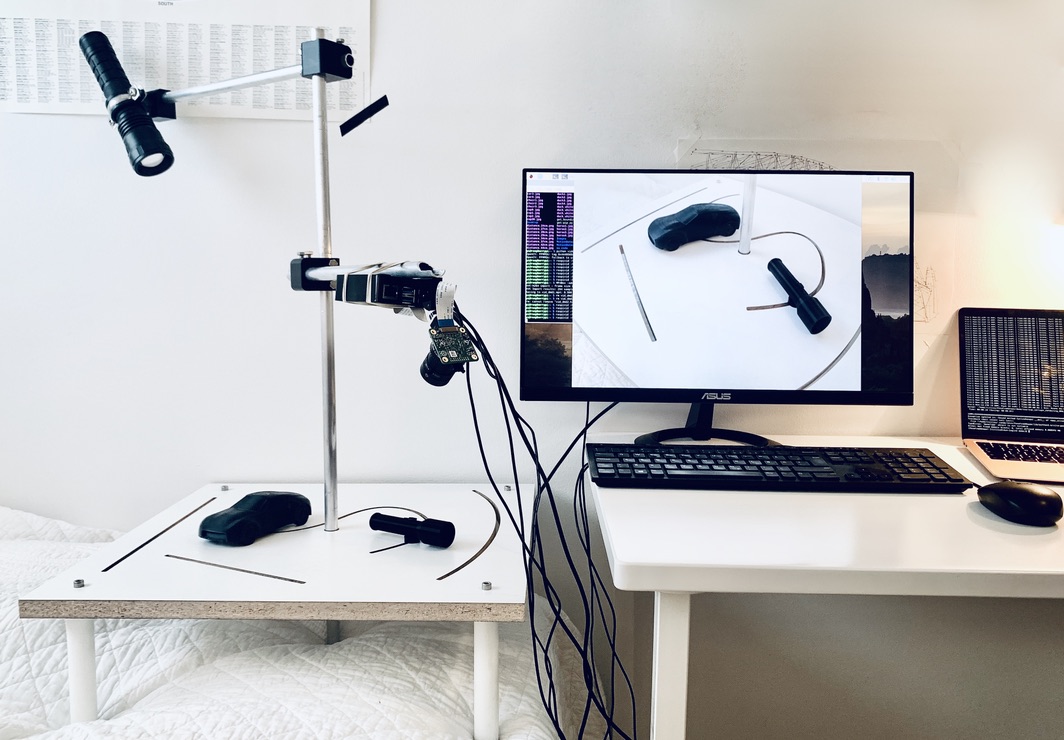}
        \caption{The Raspberry Pi 4B and Raspberry Pi camera mounted on the camera tripod}
        \label{fig:experimental_camera_closeup}
    \end{figure}
    The experimental setup (shown in \autoref{fig:experimental_camera_closeup}) developed for this work consists of a square platform with laser cut slots along which the 3D models can move and rotate. A vertical rod at the center supports two more steel arms capable of rotating along the axis defined by the vertical rod. One of these arms support a light source while the other supports the Raspberry Pi camera setup. The Raspberry Pi 4 (\cite{gay_raspberry_2014}) is the fourth generation single-board computer developed by the Raspberry Pi Foundation in the United Kingdom. It is widely used within a range of fields. The Raspberry Pi is designed to run Linux, and it features a processing unit powerful enough to run all the algorithms utilized in this work. The hardware specifications of the Raspberry Pi 4 are presented in the \autoref{table:raspberrypi}. The Raspberry Pi 4 was connected to a Raspberry Pi HQ camera, version 1.0, and a $6mm$ compatible lens during experiments. The specification of the camera is given in \autoref{tab:raspberry_camera_specifiations}. 

    \begin{table}
    \centering
    \caption{Specification of Raspberry Pi 4 Model B}
    \label{table:raspberrypi}
        \begin{tabular}{cc}
            \toprule
            \textbf{Processor} & Broadcom BCM2711, quad-core Cortex-A72\\
            & (ARM v8) 64-bit SoC @ 1.5GHz\\
            \textbf{Memory} & 8GB LPDDR4\\
            \textbf{GPU} & Broadcom VideoCore VI @ 500 MHz\\
            \textbf{OS }& Raspbian Pi OS \\
            \bottomrule
        \end{tabular}
    \end{table}

    \begin{table}
    \centering
    \caption{Specification of the 6mm IR CCTV lens}
    \label{tab:raspberry_camera_specifiations}
    \begin{tabular}{cc}
            \toprule
            \textbf{Sensor} & Sony IMX477\\
            \textbf{Sensor Resolution} & $4056 \times 3040$ pixels\\
            \textbf{Sensor Image Area} & $6.287mm \times 4.712mm$\\
            \textbf{Pixel Size} & $1.55\mu \times 1.55\mu$\\
            \textbf{Focal Length} & $6mm$\\
            \textbf{Resolution} & 3 MegaPixel\\
            \bottomrule
        \end{tabular}
    \end{table}
    Both the arms can be rotated independently and moved up and down to illuminate and record the scene on the platform from different angles respectively. All the scripts for capturing and processing scene, and trained algorithms are installed on the Raspberry Pi setup for operational use. Furthermore, the Raspberry Pi is connected to a power source and an Ethernet cable to ensure a stable internet connection. An external monitor is also connected for displaying the results.
    
    \subsection{Datasets}
    \label{subsec:datasets}
    For the training of all the algorithms used in this work, two datasets: the ObjectNet3D \cite{xiang_objectnet3d_2016} and the Pascal3D \cite{xiang_beyond_2014} were used. These datasets provide 3D pose and shape annotations for various detection and classification tasks. The ObjectNet3D database consists of 100 object categories, 90,127 images containing 201,888 objects and 44,147 3D shapes. Objects in the images are aligned with the 3D shapes. The alignment provides both accurate 3D pose annotation and the closest 3D shape annotation for each 2D object, making the database ideal for 3D pose and shape recognition of 3D objects from 2D images. Similarly, the Pascal3D database consists of 12 image categories with more than 3,000 instances per category. 
    
\begin{figure}
        \includegraphics[width=.32\linewidth]{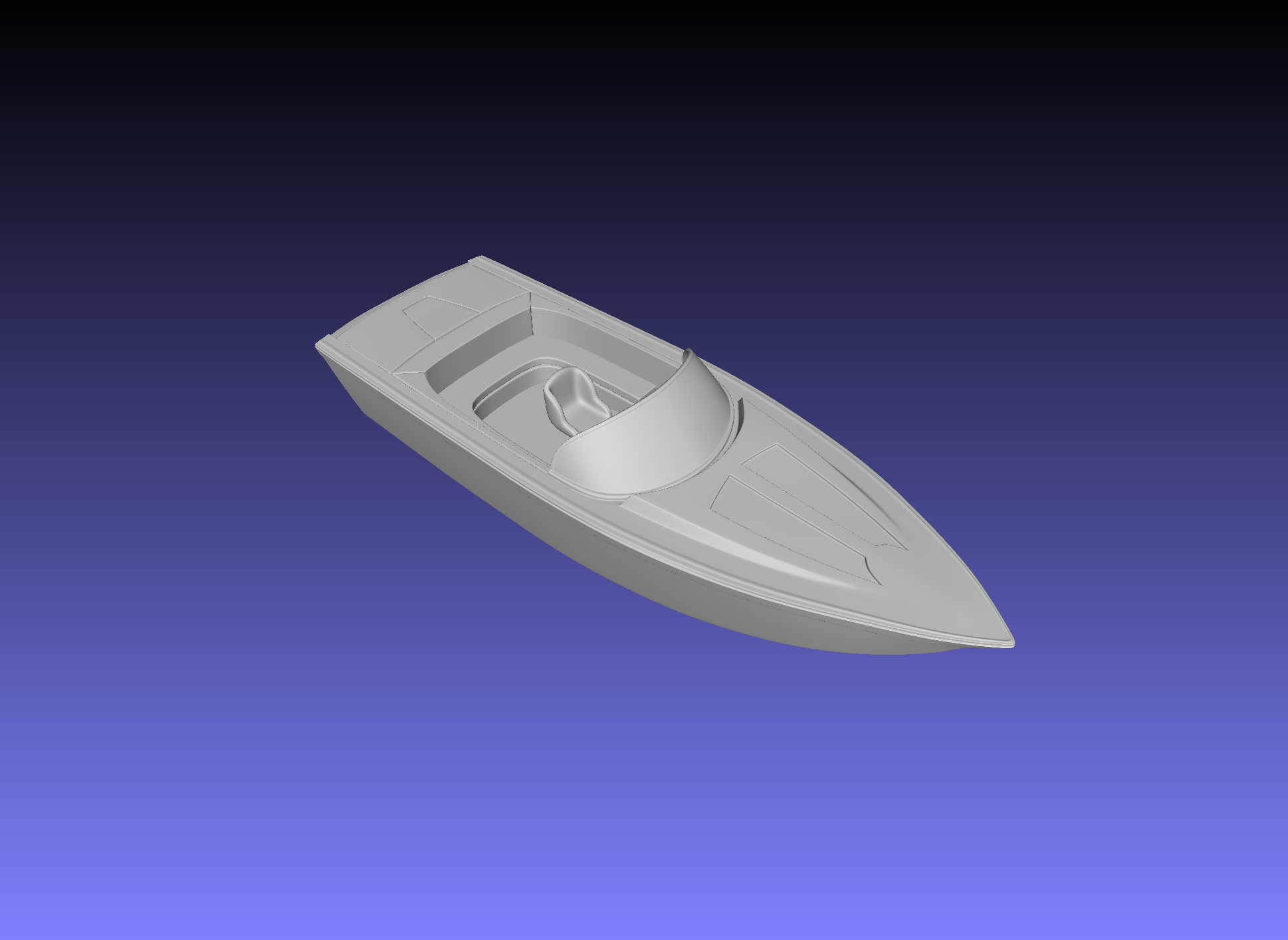}
        \includegraphics[width=.32\linewidth]{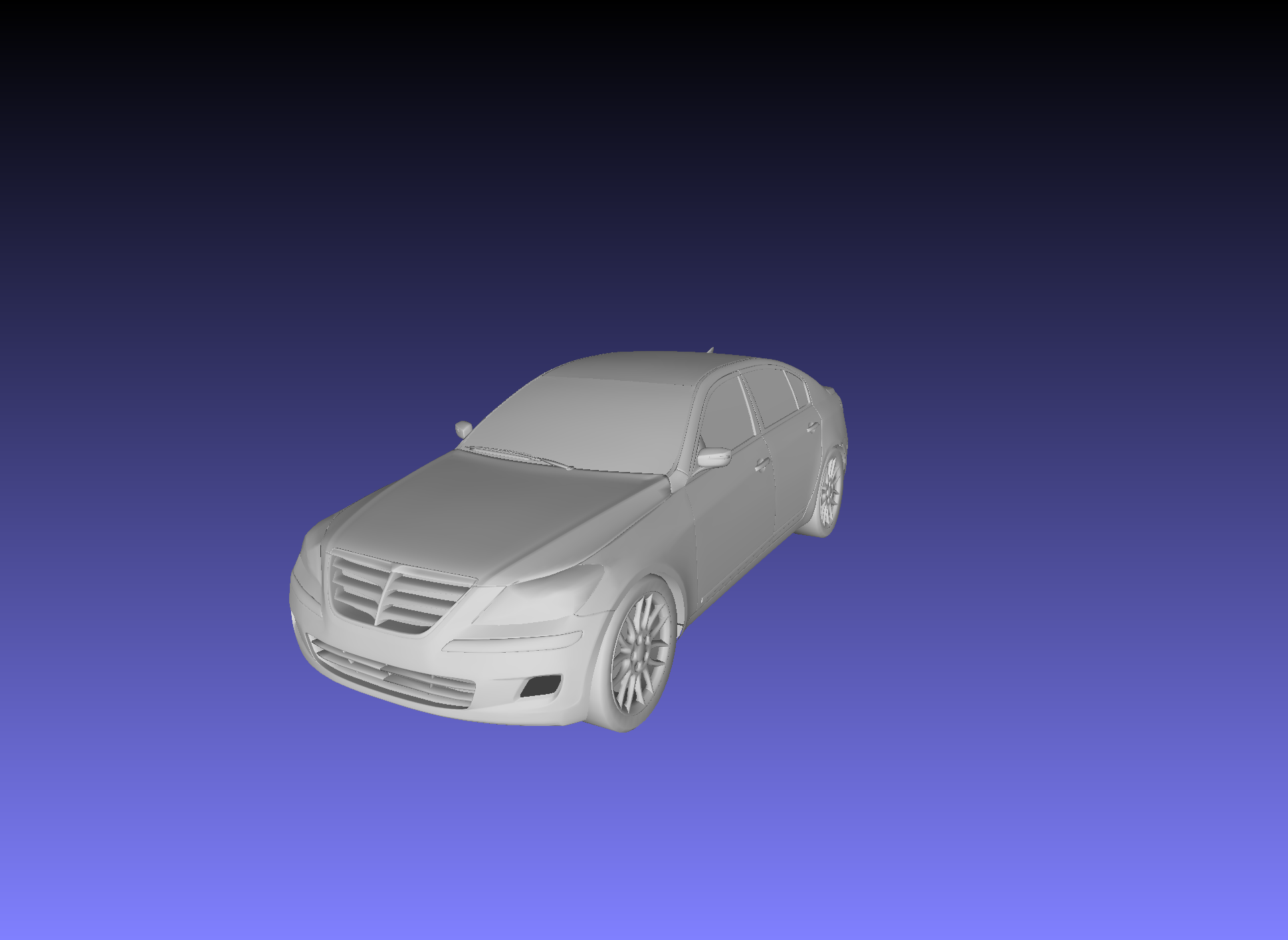}
        \includegraphics[width=.32\linewidth]{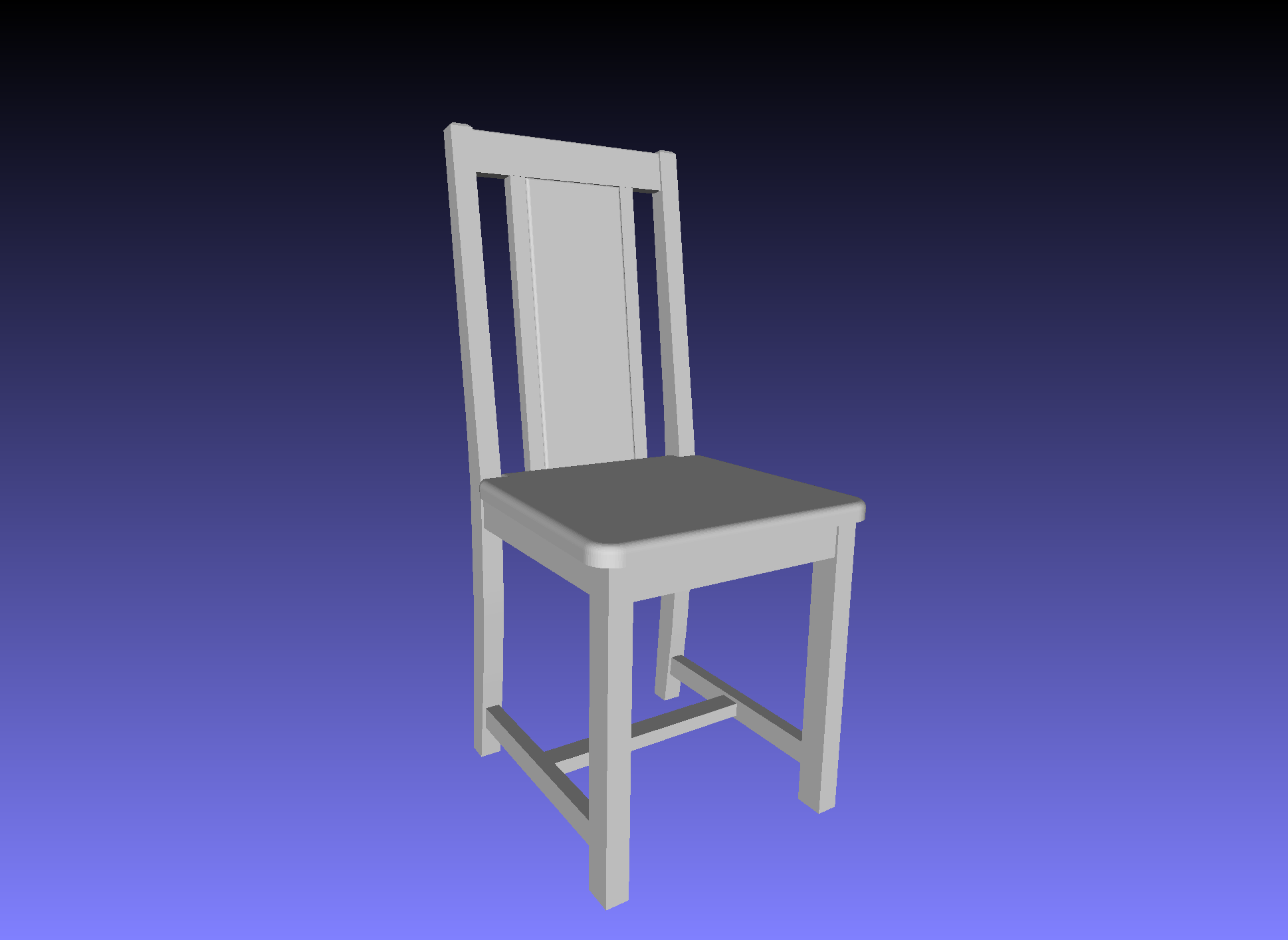}
        \\[\smallskipamount]
        \includegraphics[width=.32\linewidth]{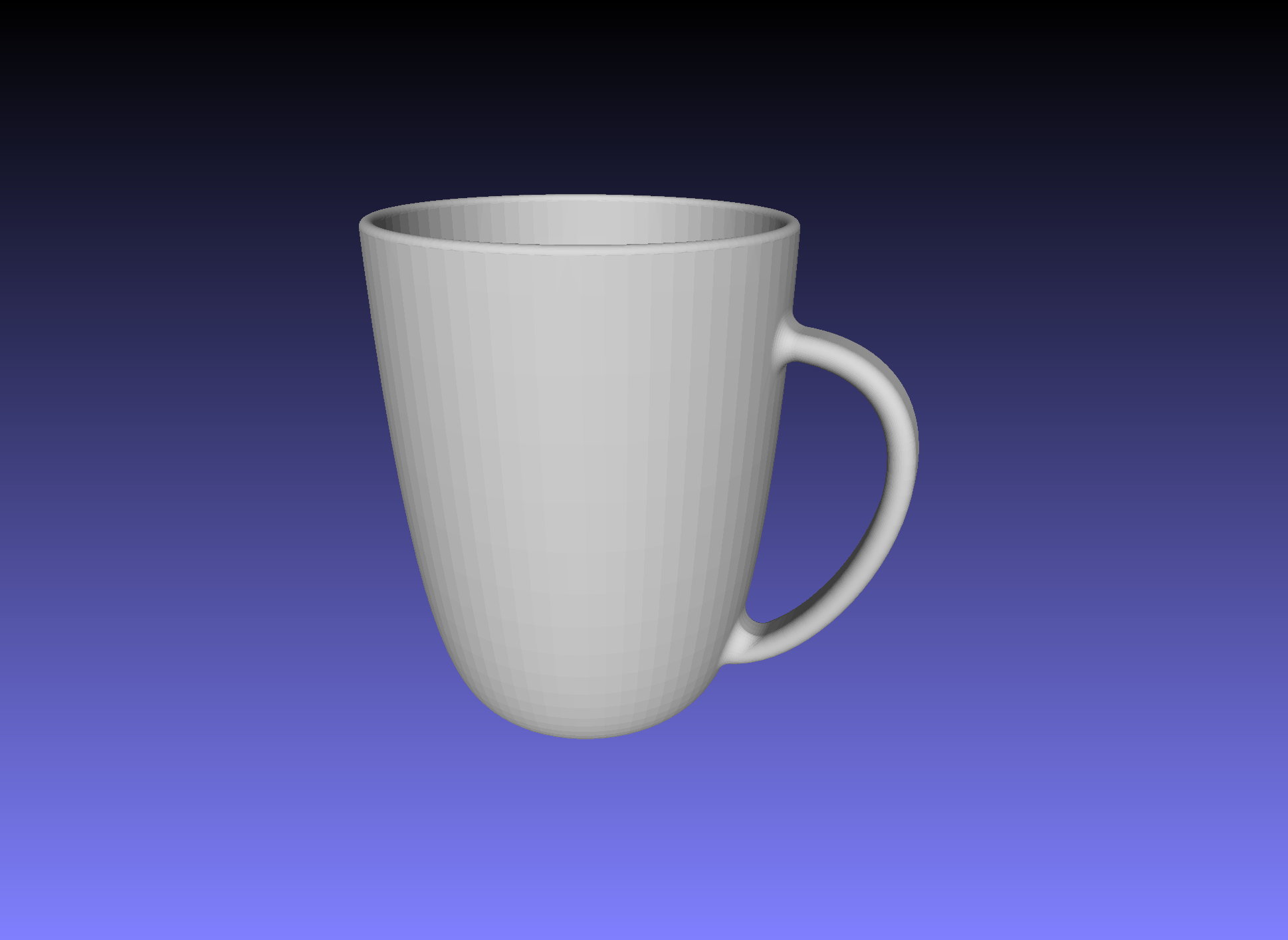}
        \includegraphics[width=.32\linewidth]{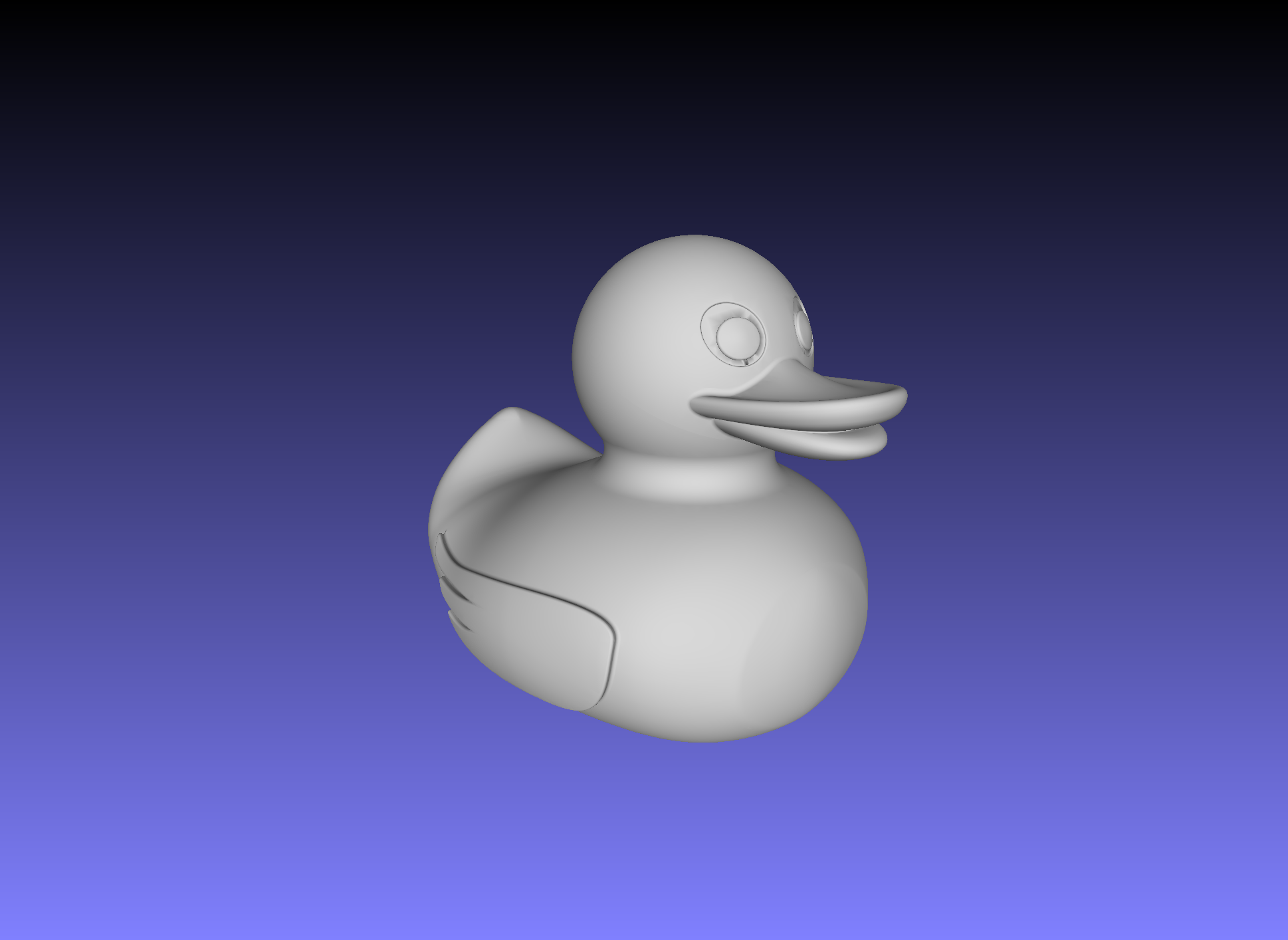}
        \includegraphics[width=.32\linewidth]{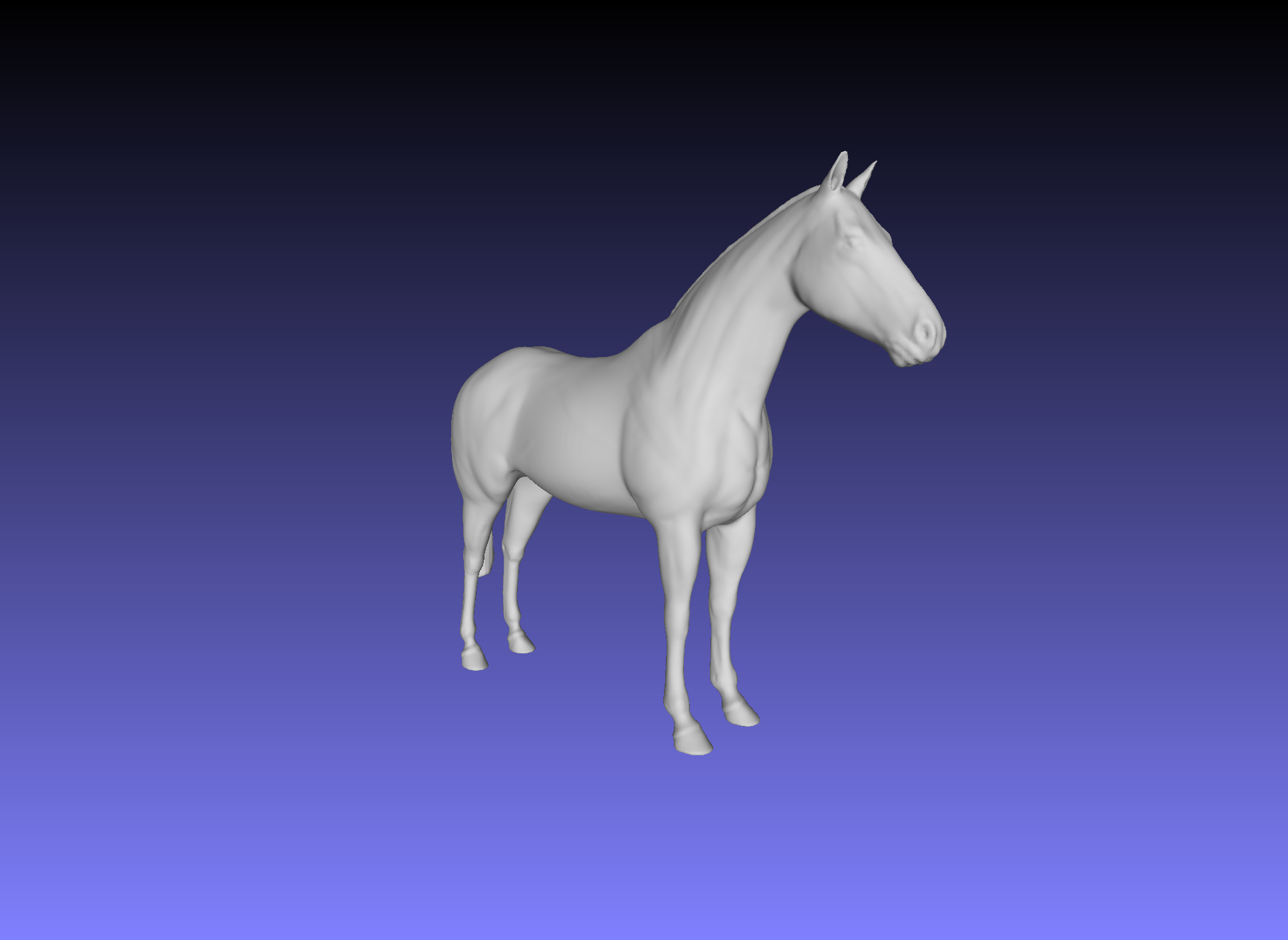}
        \\[\smallskipamount]
        \includegraphics[width=.32\linewidth]{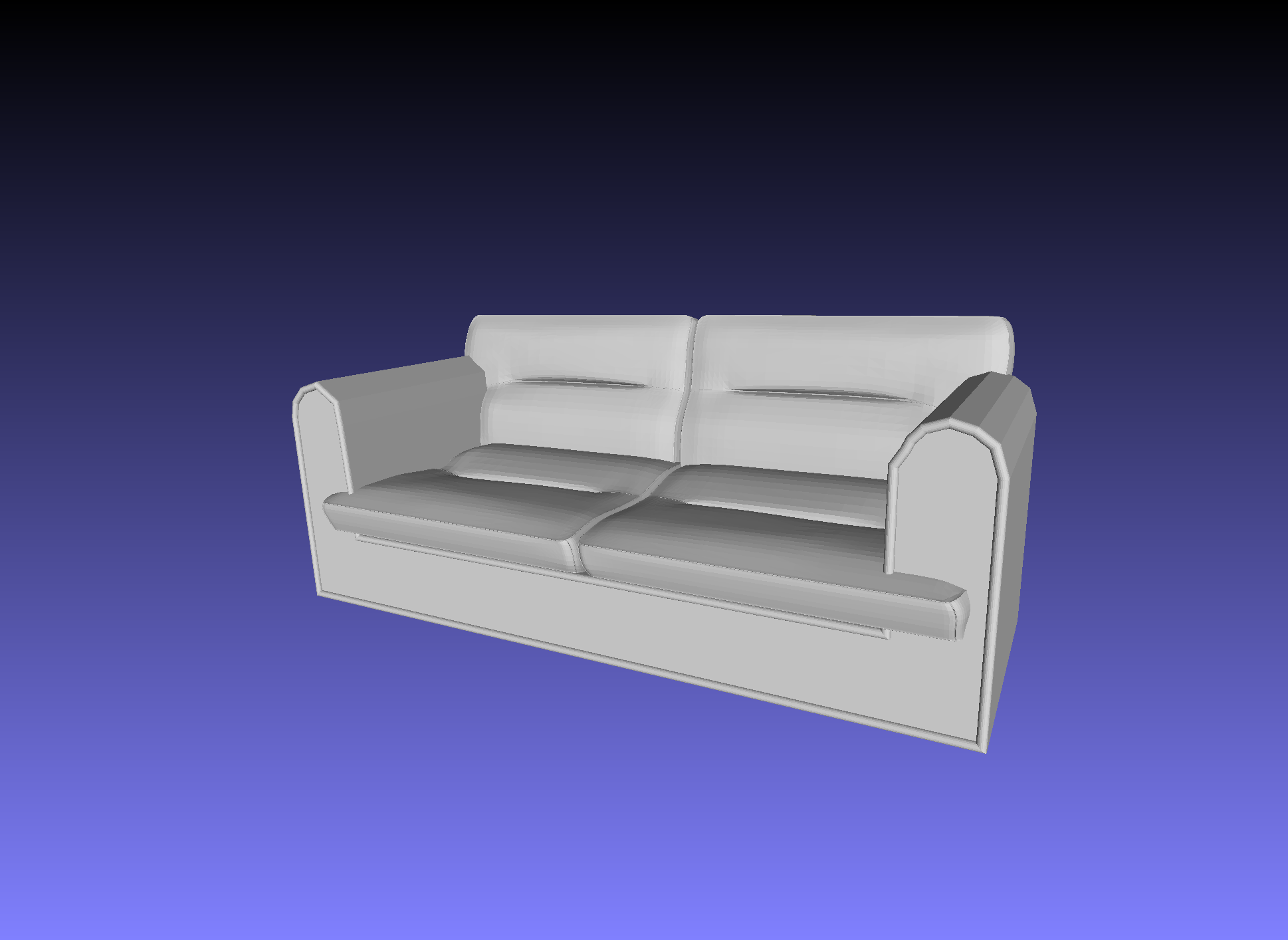}
        \includegraphics[width=.32\linewidth]{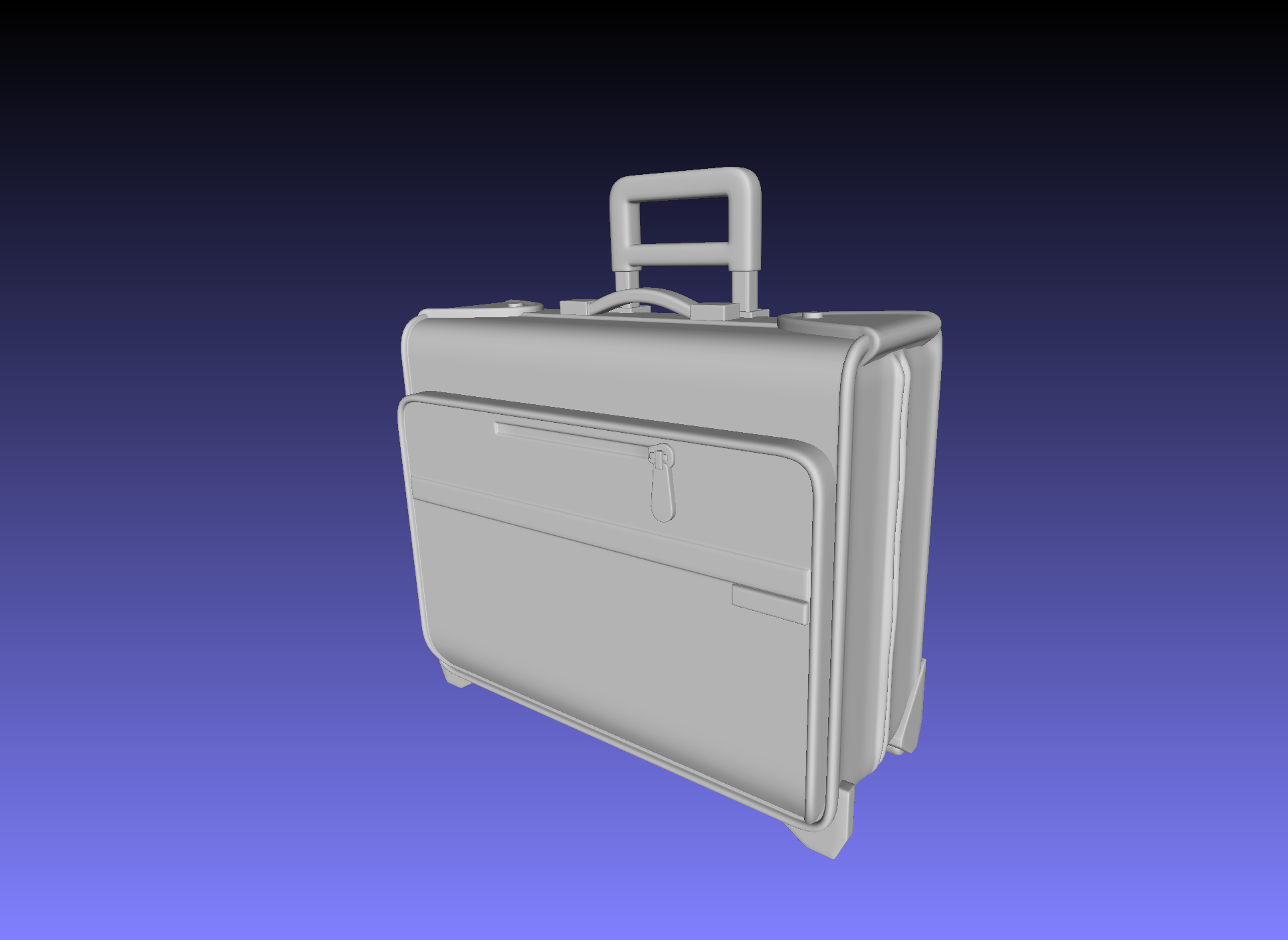}
        \includegraphics[width=.32\linewidth]{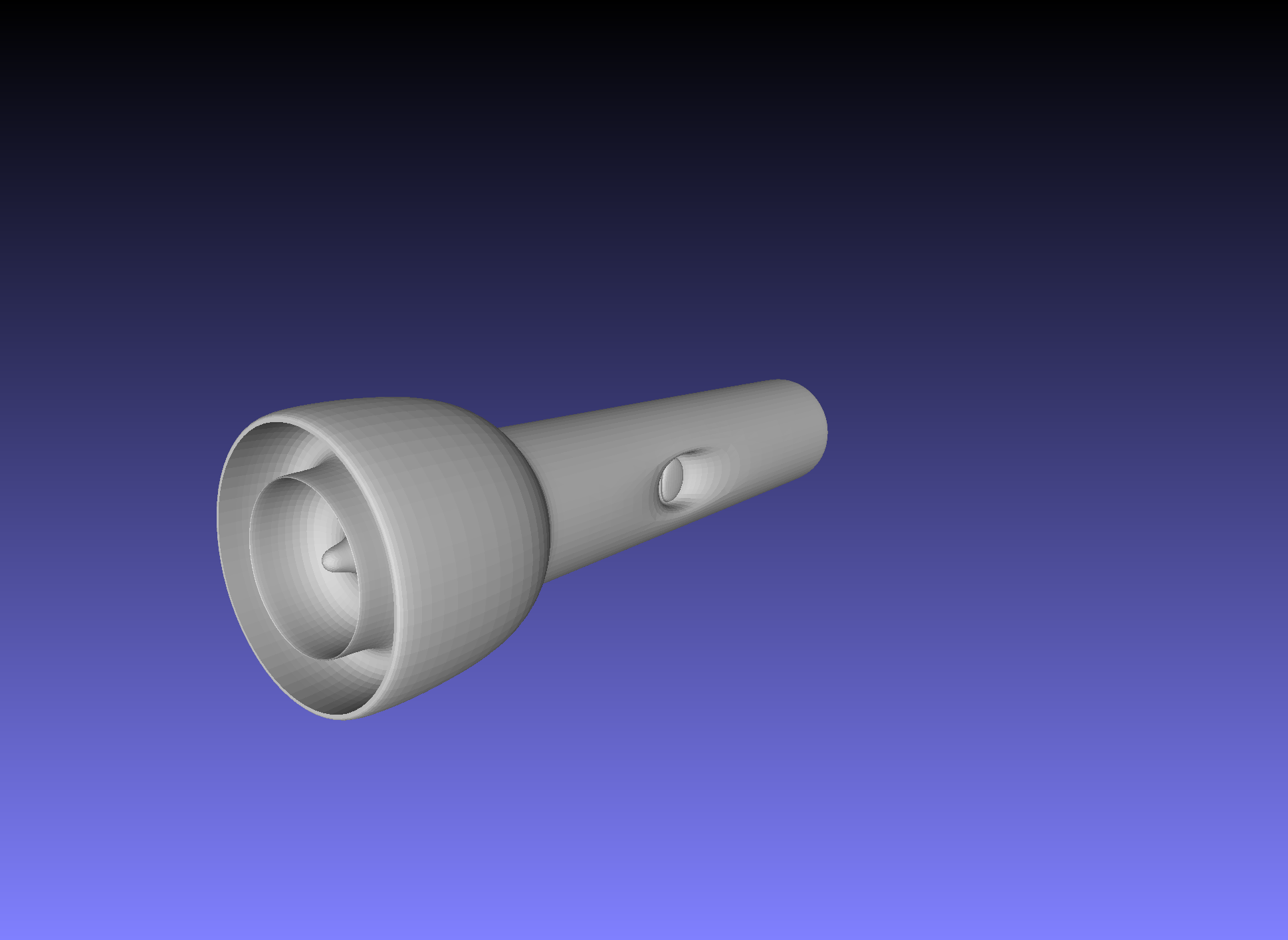}
        \caption{3D CAD models used for 3D printing and as inputs to the 3D pose estimation network, displayed in MeshLab.}
        \label{fig:3dmodels}
    \end{figure}

    The 3D CAD models (shown in \autoref{fig:3dmodels}) of the objects used in this work were download from the GrabCAD \cite{grabcad} and Free3D \cite{free3d} websites in the Standard Tessellation Language (STL) format and then converted to object (OBJ) format. The STL format stores the 3D model information as a collection of unstructured triangulated surfaces each described by the unit normal and vertices (ordered by the right-hand rule) of the triangles using a three-dimensional cartesian coordinate system. The STL files do not store any information regarding the color or texture of the surfaces. The OBJ files which is very similar to STL files also include these information. In the proposed workflow the STL files were used for 3D printing and the associated OBJ files were used as input to the pose estimation network.
    For the testing of the trained algorithms, data was collected in the form of images and videos using the  experimental set-up described in the \autoref{subsec:method_experimental_setup}. In total 450 images of the nine object categories presented in \autoref{fig:3dmodels} were collected. The images consist of single object seen from different angles, multiple objects seen from different angles and objects occluded by the other objects. In addition, the images were taken both in natural light and with additional lighting from four different angles using the torch. The image processing is further described in \autoref{subsec:method_training_YOLOv5}. 
    
    \subsection{Motion detection}
    \label{subsec:method_dmd}
    DMD is an equation-free method that is capable of retrieving intrinsic behaviour in data, even when the underlying dynamics are nonlinear (\cite{tu2013dynamic}). It is a purely data-driven technique, which is proving to be more and more important in the arising and existing age of big data. DMD decomposes time series data into spatiotemporal coherent structures by approximating the dynamics to a linear system that describes how it evolves in time. The linear operator, sometimes also referred to as the Koopman operator (\cite{nathan2017dynamic}), that describes the data from one time step to the next is defined as follows

    \begin{equation}
    \label{eq:DMD1}
        \mathbf{x}_{t+1} = \mathbf{Ax}_{t}
    \end{equation}
    
    Consider a set of sampled snapshots from the time series data. Each snapshot is vectored and structured as a column vector with dimensions $n \times 1$ in the following two matrices
    
     \begin{equation}
     \label{eq:DataMatrices}
      \mathbf{X} = \{\mathbf{x}_{1},...,\mathbf{x}_{m-1}\} , \quad \mathbf{X'} = \{\mathbf{x}_{2},...,\mathbf{x}_{m}\}
      \end{equation}
    
    Where $\mathbf{X}'$ is the $\mathbf{X}$-matrix shifted one time step ahead, each of dimension $n \times (m-1)$.
    
    Relating each snapshot to the next, \autoref{eq:DMD1} can be rewritten more compactly as
    
    \begin{equation}
    \label{eq:DMD2}
        \mathbf{X^{'}} = \mathbf{AX}
    \end{equation}
    
    The objective of DMD is to find an estimate of the linear operator $\mathbf{A}$ and obtain its leading eigenvectors and eigenvalues. This will result, respectively, in the modes and frequencies that describes the dynamics.
    
    Computing the leading DMD modes of the linear operator proceeds as follows
    
    \textbf{Algorithm 1:} Standard DMD
    \begin{enumerate}
     \item Structure data vectors into matrices $\mathbf{X}$ and $\mathbf{X'}$ as described in \autoref{eq:DataMatrices}
      \item Compute the SVD of $\mathbf{X}$
      \begin{align}
          \mathbf{X} = \mathbf{U}\boldsymbol{\Sigma}\mathbf{V^*}
      \end{align}
            Where $\mathbf{U}$ and $\mathbf{V}$ are square unitary matrices of sizes $n \times n$ and $m \times m$ respectively, and $\mathbf{UU^{*}} = \mathbf{VV^{*}} = \mathbf{I}$ .
    \item To reduce the order of the system, the following matrix is defined
        \begin{align}
          \mathbf{\Tilde{A}} = \mathbf{U^{*}AU} = \mathbf{U^{*}X'\mathbf{V}}\boldsymbol{\Sigma{}^{-1}}
      \end{align}
            Where $\mathbf{\Tilde{A}}$ is projected onto the $r$ leading modes of $\mathbf{U}$ as a result of a truncated SVD. 
        \item Compute the eigendecomposition of $\mathbf{\Tilde{A}}$:
        \begin{align}
         \mathbf{\Tilde{A}W} = \mathbf{W}\boldsymbol{\Lambda}
      \end{align}
      \item Which ultimately leads to the DMD \emph{modes} $\boldsymbol{\Psi}$, where each column of $\mathbf{W}$ represents a single mode of the solution.
      \begin{align}
         \boldsymbol{\Psi} = \mathbf{X^{'}V}\boldsymbol{\Sigma}^{-1}\mathbf{W}
      \end{align}
    \end{enumerate}
    The initial amplitude of the system is obtained by solving $\boldsymbol{\Psi b}_{0} = \mathbf{x}_{0}$.
    The predicted state is then expressed as a linear combination of the identified modes.
    
    \begin{equation}
        \hat{\mathbf{x}} = \boldsymbol{\Psi\Lambda}^{k}\mathbf{b}_{0}
    \end{equation}
    
    Note that each DMD mode $\boldsymbol{\psi}_{j}$ is a vector that contains the spatial information of the decomposition, while each corresponding eigenvalue $\lambda_{j}^{k}$ along the diagonal of $\boldsymbol{\Lambda}^{k}$ describes the time evolution. The part of the video frame that changes very slowly in time must therefore have a stationary eigenvalue, i.e. $\mid\lambda\mid \approx 1$. This fact can be used to separate the slowly varying background section in a video from the moving foreground elements, hence enabling object localisation.
    
    \subsection{Object detection and classification}
    \label{subsec:method_training_YOLOv5}
    For detecting objects in images, we chose the YOLOv5, which is the latest version in the family of YOLO algorithms at the time of writing. YOLOv5 is the first YOLO implementation written in the PyTorch framework, and it is therefore considered to be more lightweight than previous versions while at the same time offering great computational speed. There are no considerable architectural changes in YOLOv5 compared to the previous versions YOLOv3 and YOLOv4. In the current work YOLOv5 was retrained to recognize the 3D CAD models from our own experimental set-up as specified in \autoref{subsec:datasets}. We used Computer Vision Annotation Tool (CVAT) for labeling images. The CVAT is an open-source, web-based image annotation tool produced by Intel. To annotate the images, we drew bounding boxes around the objects that we wanted our detector to localize, and assigned the correct object categories. The most important hyperparameters for the YOLO algorithm is presented in \autoref{tab:yolo_parameters}.
        \begin{table}
        \centering
        \caption{Choice of training parameters for YOLOv5x}
        \label{tab:yolo_parameters}
        \begin{tabular}{cc}
        \toprule
        \textbf{Parameter} & \textbf{Value}\\
        \midrule
        Image size & 640 \\
        Batch size & 16 \\
        Epochs & 300 \\
        Device & GPU \\
        Learning rate & $10^{-3}$\\
        \bottomrule
        \end{tabular}
    \end{table}
    
    \begin{figure}
    \begin{subfigure}[b]{\linewidth}
        \includegraphics[width=\linewidth]{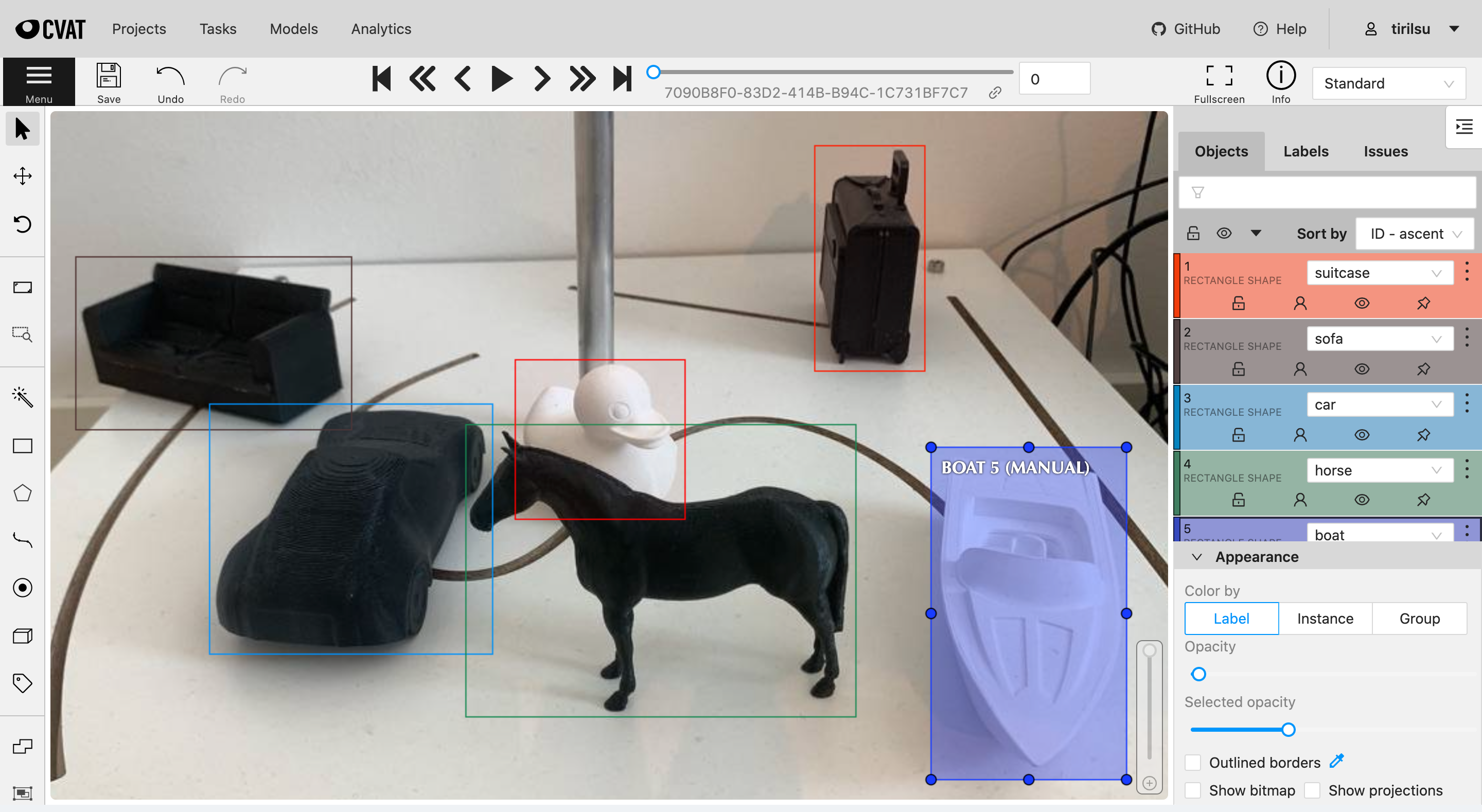}
       \caption{Screenshot of the image annotation process in CVAT}
    \end{subfigure}
    \newline
    \newline
    \begin{subfigure}[b]{\linewidth}
        \includegraphics[width=\linewidth]{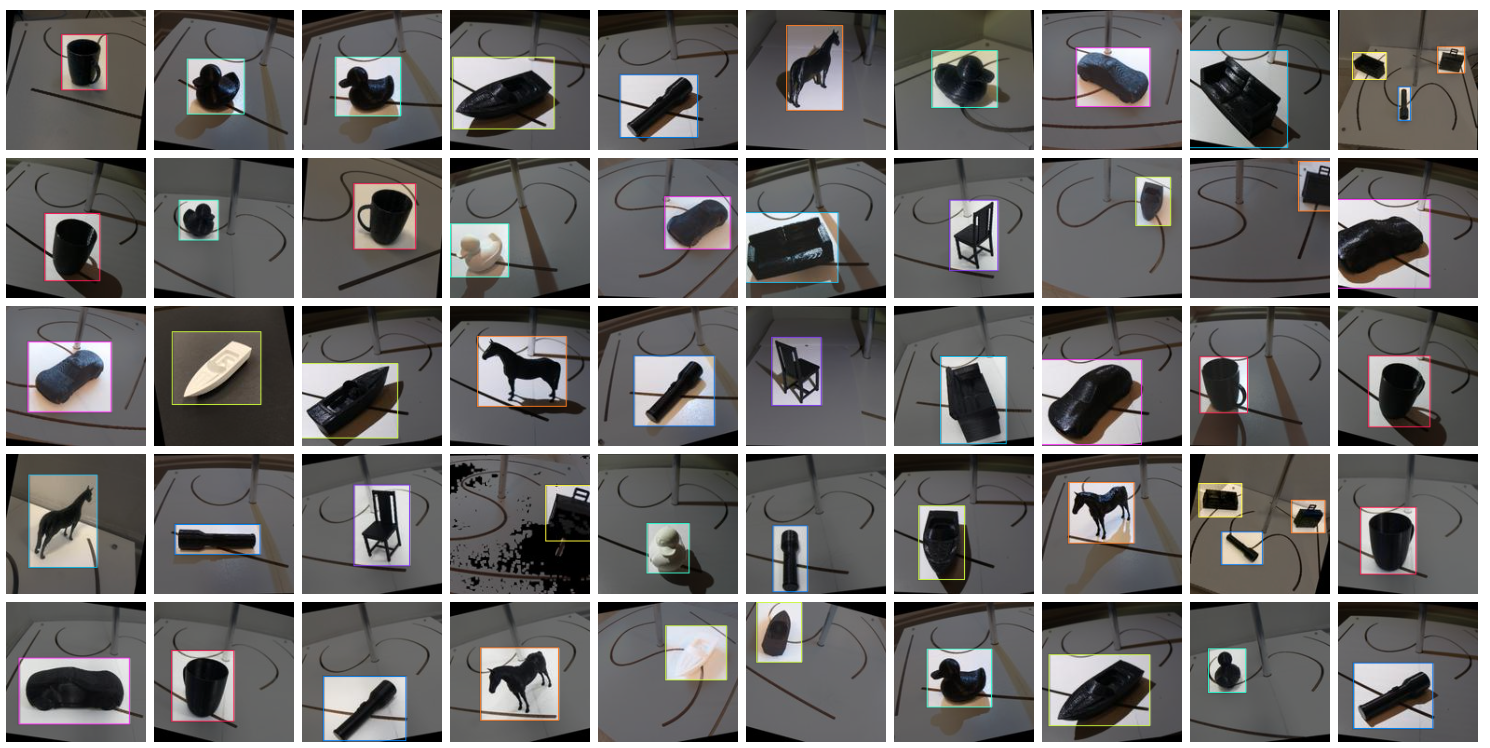}
        \caption{Screenshot of the fully annotated dataset in Roboflow}
    \end{subfigure}
    \caption{Procedure of annotating dataset used for supervised learning in image annotation tools CVAT and Roboflow}
    \label{fig:data_annotation}
    \end{figure}
    
    Furthermore, the fully annotated dataset was opened in Roboflow where techniques of preprocessing and augmentation were applied. Data augmentation involved  rotation, cropping, gray-scaling, changing exposure, and adding noise. The final image dataset was split into training, validation and test sets. The respective ratios corresponded to 70:20:10. The procedures from CVAT and Roboflow are illustrated in \autoref{fig:data_annotation}.
    
     \subsection{Pose estimation}
    \label{subsec:poseestimation}
    For pose estimation, we implemented the method presented by \citet{xiao_pose_2019}. The essence of the method is to combine image and shape information in the form of 3D models to estimate the pose of a depicted object. The information that relates a depicted object to its 3D shape stems from the feature extraction part of the method. Two separate feature extractions are performed in parallel. One for the images themselves by using a CNN (ResNet-18), and one for the corresponding 3D shapes. The 3D shape features are extracted either by feeding the 3D models as point clouds to the point set embedding network PointNet \cite{qi2017pointnet}, or by representing the 3D shape through a set of rendered image views that circumvents the object at different orientations, and further passes these images into a CNN.
    
    The outputs of the two feature extraction branches are then concatenated into a global feature vector prior to initiating the pose estimation part. This part consists of a fully connected multi-layer perceptron (MLP) with three hidden layers of size 800-400-200, where each layer is followed by a batch normalization and a ReLU activation. The network outputs the three Euler angles of the camera; azimuth, elevation and in-plane rotation, with respect to the shape reference frame. Each angle is estimated as belonging to a certain angular bin $l \in \{0,L_{\theta} - 1\}$, for a uniform split of $L_{\theta}$ bins. This is done through a softmax activation in the output layer which yields the bin-probabilities. Along with each predicted angle belonging to an angular bin, an angle offset, $\hat{\delta}_{\theta, l} \in [-1, 1]$  relative to the center of an angle bin is estimated to obtain an exact predicted angle.
    
\begin{figure}
        \includegraphics[width=.19\linewidth]{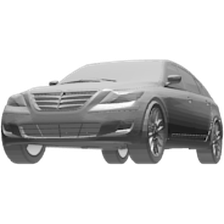}\hfill
        \includegraphics[width=.19\linewidth]{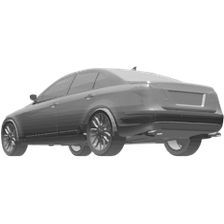}\hfill
        \includegraphics[width=.19\linewidth]{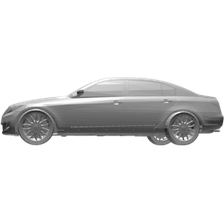}\hfill
        \includegraphics[width=.19\linewidth]{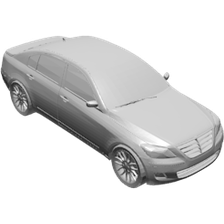}
        \includegraphics[width=.19\linewidth]{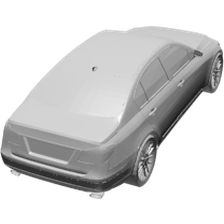}
        \caption{A selection of rendered images using Blender}
        \label{fig:PoseFromShape_render_utils}
    \end{figure}

    For this report, the 3D model features are extracted by using the method of rendering different views of the 3D shape as mentioned above, and using them as input to a CNN. This is done with the help of the Blender module for Python \cite{noauthor_blender_2018}.  A total of 216 images are rendered per input object. An illustration of some rendered images that were collected with Blender are depicted in \autoref{fig:PoseFromShape_render_utils}. 
    
\begin{figure}
        \includegraphics[width=\linewidth]{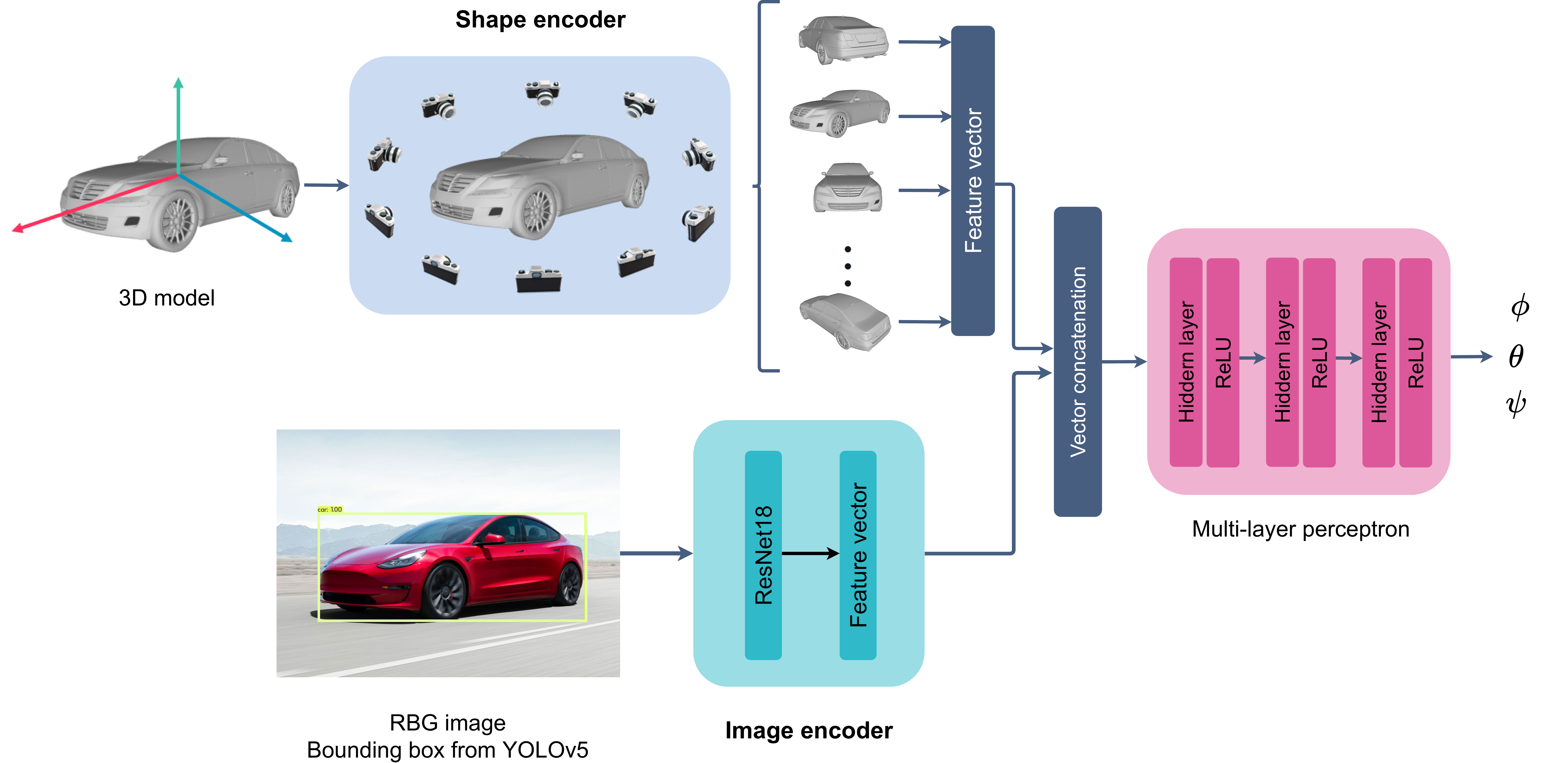}
        \caption{Architecture of the 3D pose estimation network implemented in Xiap et al. [66]}
        \label{fig:pose_estimation_workflow}
    \end{figure}

    For the image features, YOLOv5 is first applied to create bounding boxes to localize each object, before being fed to the ResNet-18. The full workflow is illustrated in \autoref{fig:pose_estimation_workflow}.
    
    The final MLP network was trained on the ObjectNet3D dataset, as this provided better system performance than the network trained on Pascal3D. Furthermore, the network was trained using the Adam optimizer for 300 epochs. The parameters used for training this part of the network are presented in \autoref{tab:pfs_parameters}.
    
    By comparing the real change in rotation of our camera with the estimated rotation angles, we get the pose estimation error. The angle output from a 3D pose estimation may be defined using a rotation matrix, eventually describing an object's change in pose.
    
     \begin{table}
     \centering
     \caption{Choice of training parameters for the 3D pose estimation network}
        \label{tab:pfs_parameters}
        \begin{tabular}{cc}
        \toprule
        \textbf{Parameter} & \textbf{Value}\\
        \midrule
        Batch size & 16 \\
        Epochs & 300 \\
        Device & GPU \\
        Learning rate & $10^{-4} / 10^{-5}$ \\
        \bottomrule
        \end{tabular}
    \end{table}
    
     \subsection{Software and hardware frameworks}
    Both the object detection and 3D pose estimation architectures used in this work are implemented in Python 3.6 using the open-source machine learning library PyTorch \cite{paszke_pytorch_2019}. The pose estimation network uses Blender 2.77, an open-source library implemented as a Python module to render multiviews of the 3D CAD models. Furthermore, the MeshLab software is used to visualize 3D objects, and the Python library Matplotlib is used for creating most of the plots and diagrams in the article. The DMD algorithm is also implemented in Python 3.6, enabling real-time processing through the OpenCV (Open Source Computer Vision Library) \cite{bradski_opencv_2000} which is an open-source software library for computer vision and machine learning.
        
    The pose estimation architecture in this project was trained and tested on the NTNU Idun computing cluster (\cite{sjalander_epic_2020}). The cluster has more than 70 nodes and 90 GPGPUs. Each node contains two Intel Xeon cores and a minimum 128 GB of main memory. All the nodes are connected to an Infiniband network. Half of the nodes are equipped with two or more Nvidia Tesla P100 or V100 GPUs. Idun's storage consists of two storage arrays and a Lustre parallel distributed file system. The object detection network applied in this work is trained and tested using Google Colab. Google Colab is an open-source computing platform developed by Google Research. The platform is a supervised Jupyter notebook service where all processing is performed in the cloud with free access to GPU computing resources. Google Colab provides easy implementations and compatibility between different machine learning frameworks. Once all the algorithms are trained they are installed on the Rasp Berry Pi for operational use.

\section{Results and Discussion}
\label{sec:resultsanddiscussion}
This section presents the results obtained from the different steps of our proposed geometric change detection workflow. Video frames where DMD detects the end of motion are used as input to the object detection network. Here, a pre-trained YOLOv5 network detects objects in the images and crops the images according to their bounding box estimations. The cropped images are then used as inputs to the pose estimation network, together with the respective CAD models of the objects.

We now present the results of motion detection using DMD followed by the results obtained from the object detection using YOLO. Lastly, we present the results from the pose estimation using singe RGB image. Throughout this section we discuss the applicability of the work to digital twins.

    \subsection{Motion Detection with DMD}
    \label{subsec:results_dmd}
    The first step in our workflow, as earlier mentioned, involves real-time motion detection and identification of moving objects in the videos recorded under different conditions. For this article six videos were recorded using the experimental setup in an operational mode. Three of these videos were recorded under normal conditions without any disturbances. These are therefore, referred to as the baseline videos. The other three videos were recorded under the influence of three distinct kind of disturbances which are considered common challenges in the context of digital twin. The three challenges whose impact on the motion detection is evaluated are as follows:
    
    \begin{itemize}
        \item \textbf{Dynamic background:} This includes objects moving in the background without being a part of the digital twin.
        \item \textbf{Lighting conditions:} This includes poor lighting condition that fails to illuminate the solid objects resulting in objects in the scene appearing like blobs. The problem is augmented for solid objects of very dark colors. In addition, the camera-lens system uses a few seconds to calibrate the lighting conditions at the beginning of the video sequences. This sudden change in the lighting condition (exposure, contrast and brightness) can be mistaken for motion.   
        \item \textbf{Monochrome frames}: This includes situation where the moving object have the similar color and pixel intensity as the background.
    \end{itemize}
    
    \begin{figure}
        \includegraphics[width=.32\linewidth]{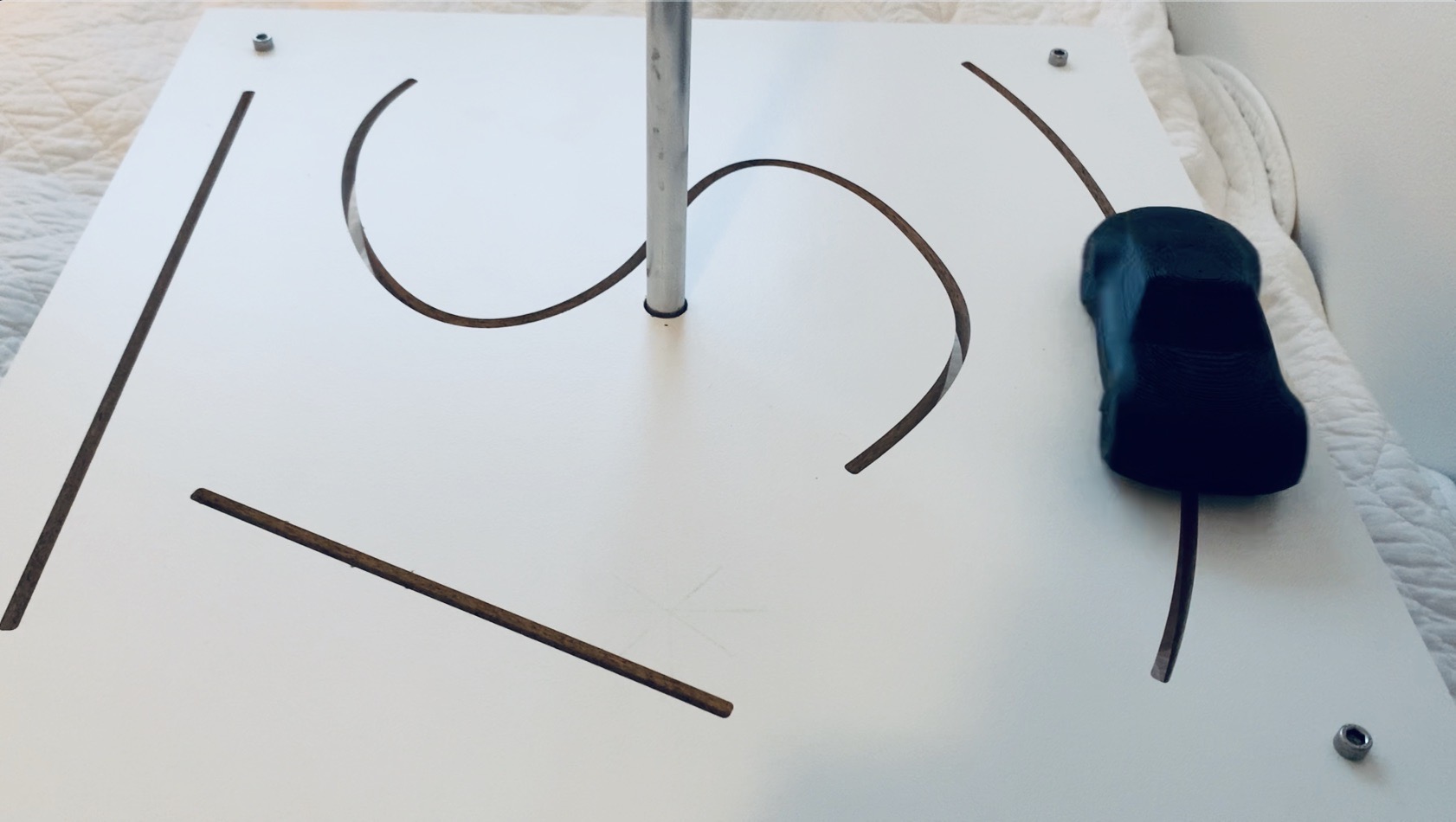}\hfill
        \includegraphics[width=.32\linewidth]{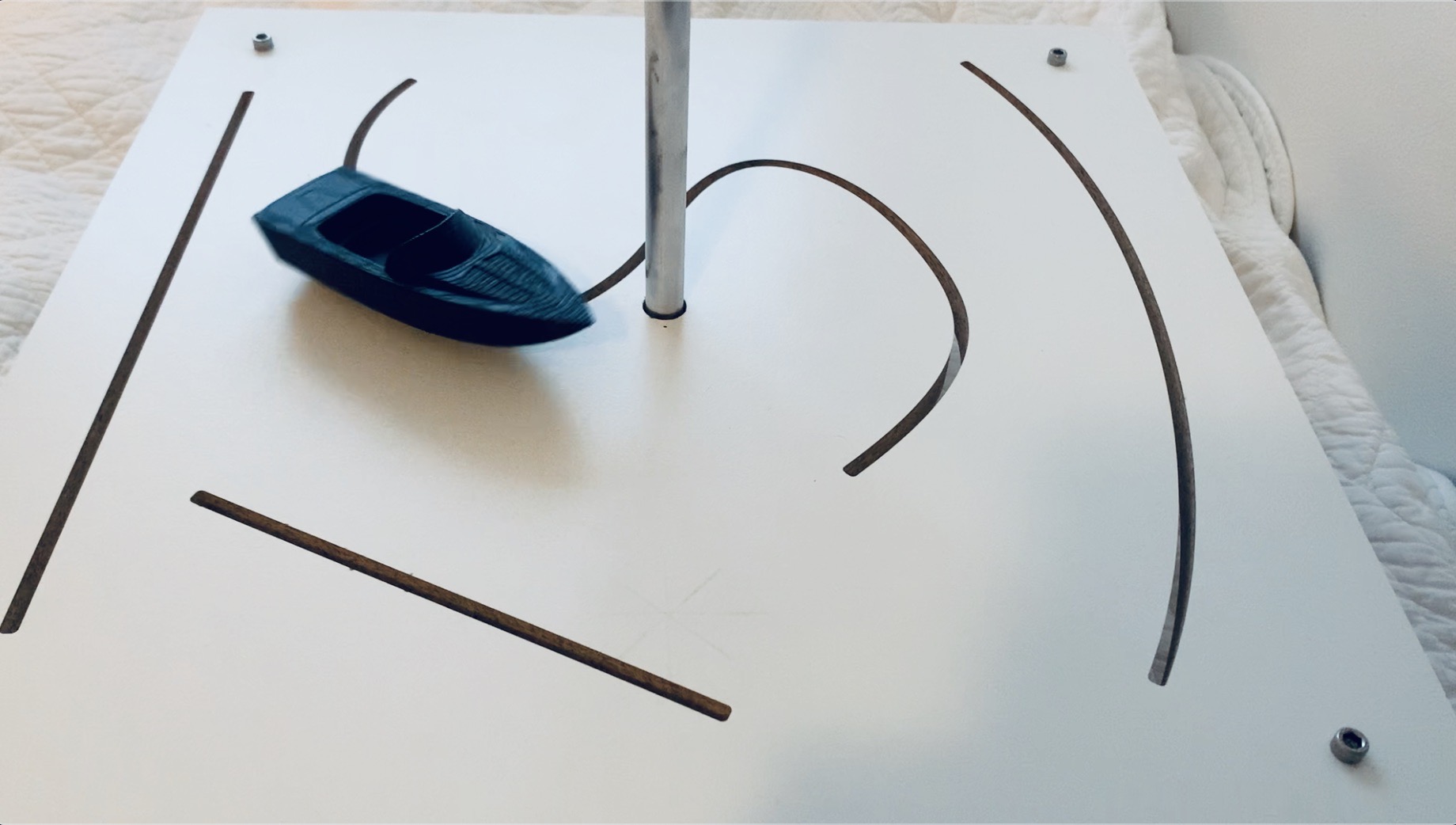}\hfill
        \includegraphics[width=.32\linewidth]{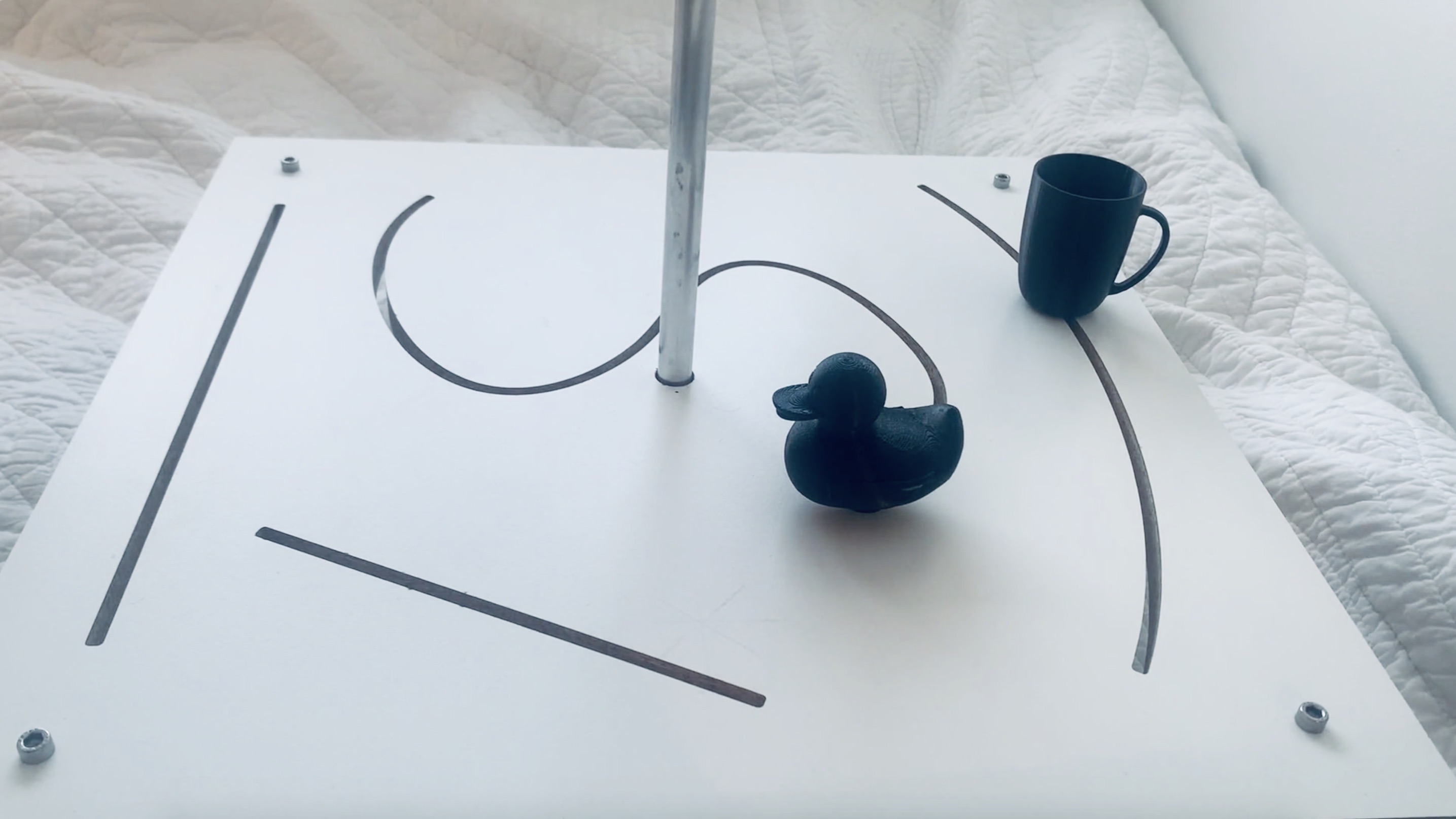}
        \\[\smallskipamount]
        \includegraphics[width=.32\linewidth]{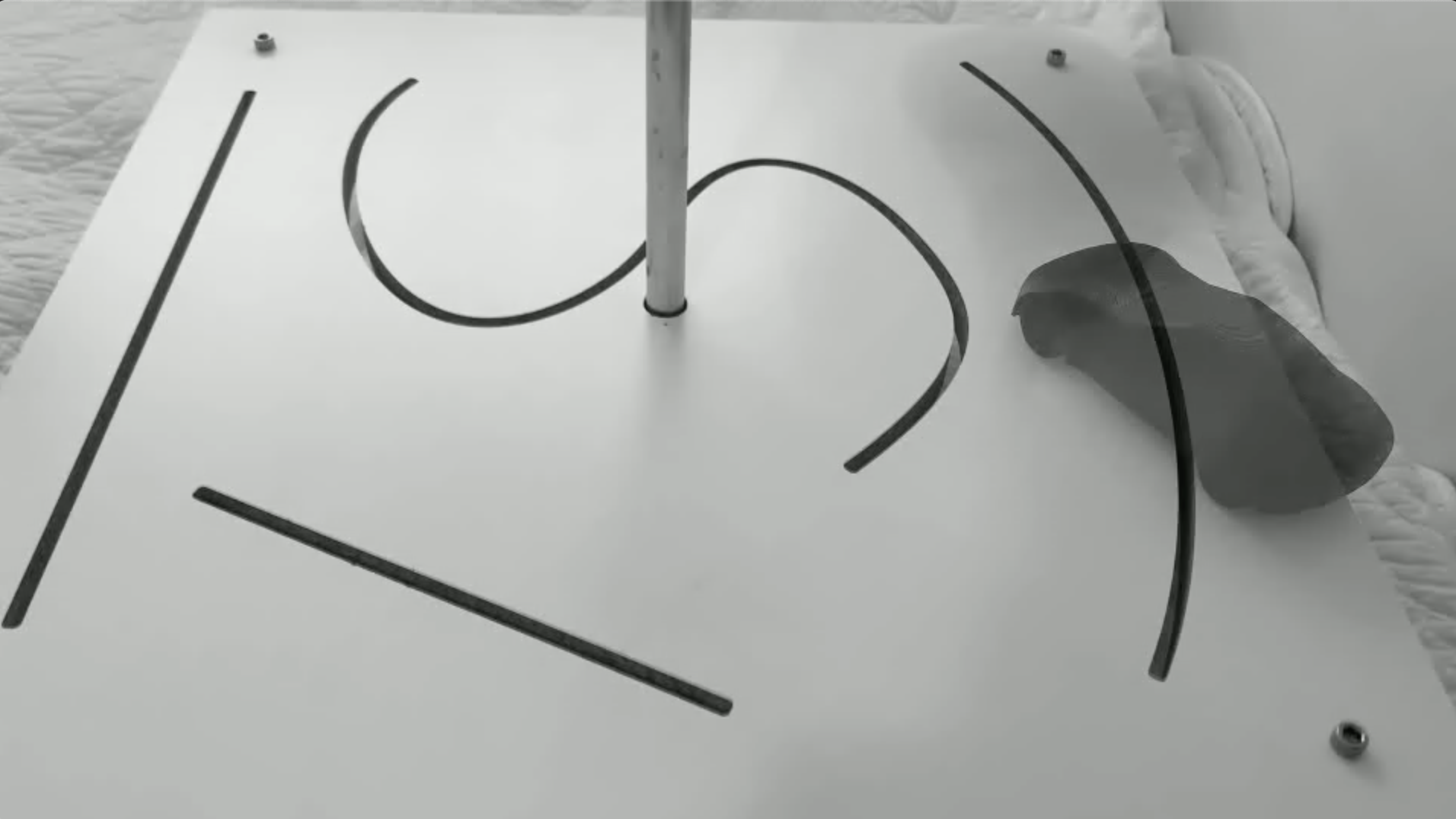}\hfill
        \includegraphics[width=.32\linewidth]{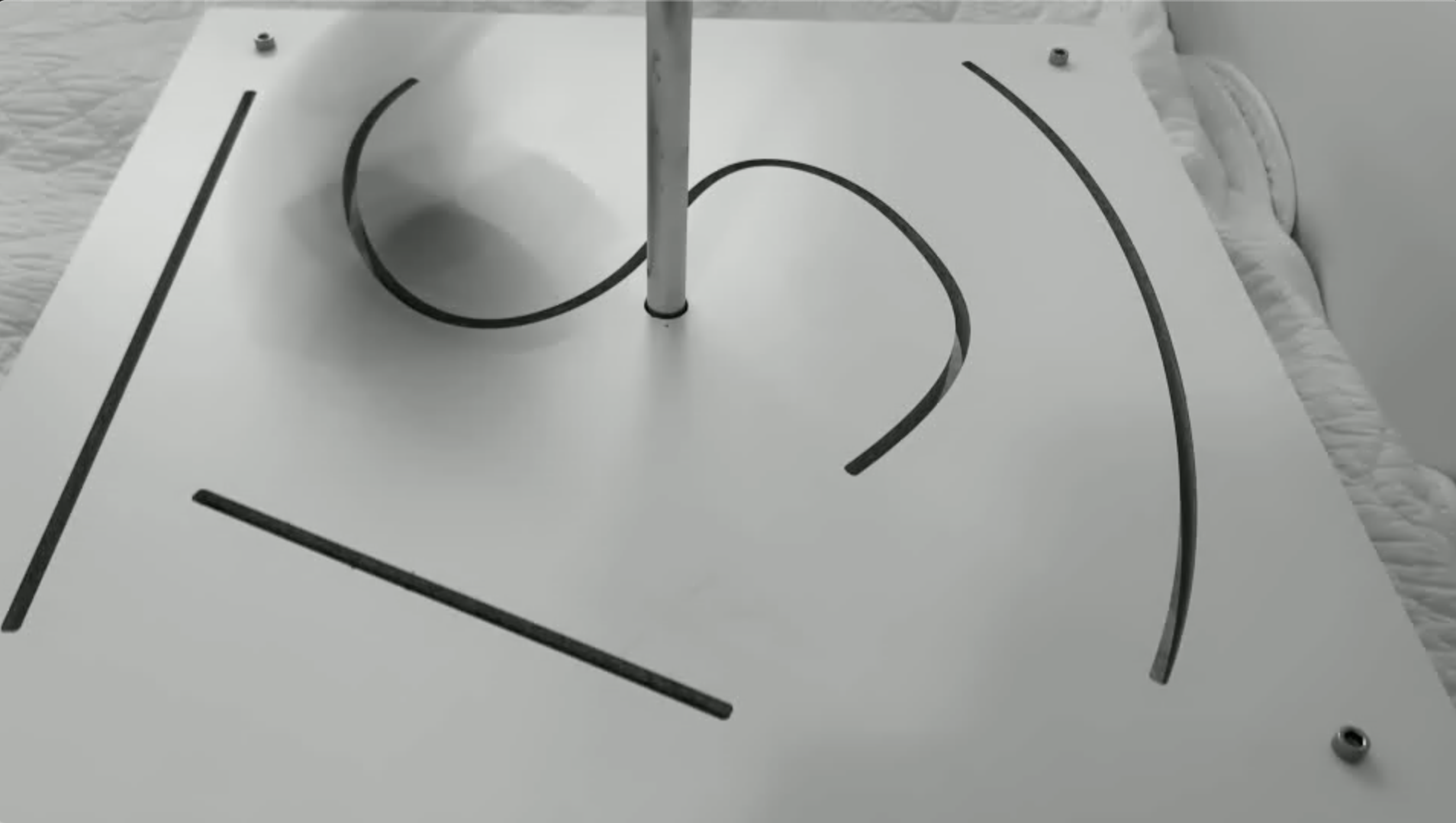}\hfill
        \includegraphics[width=.32\linewidth]{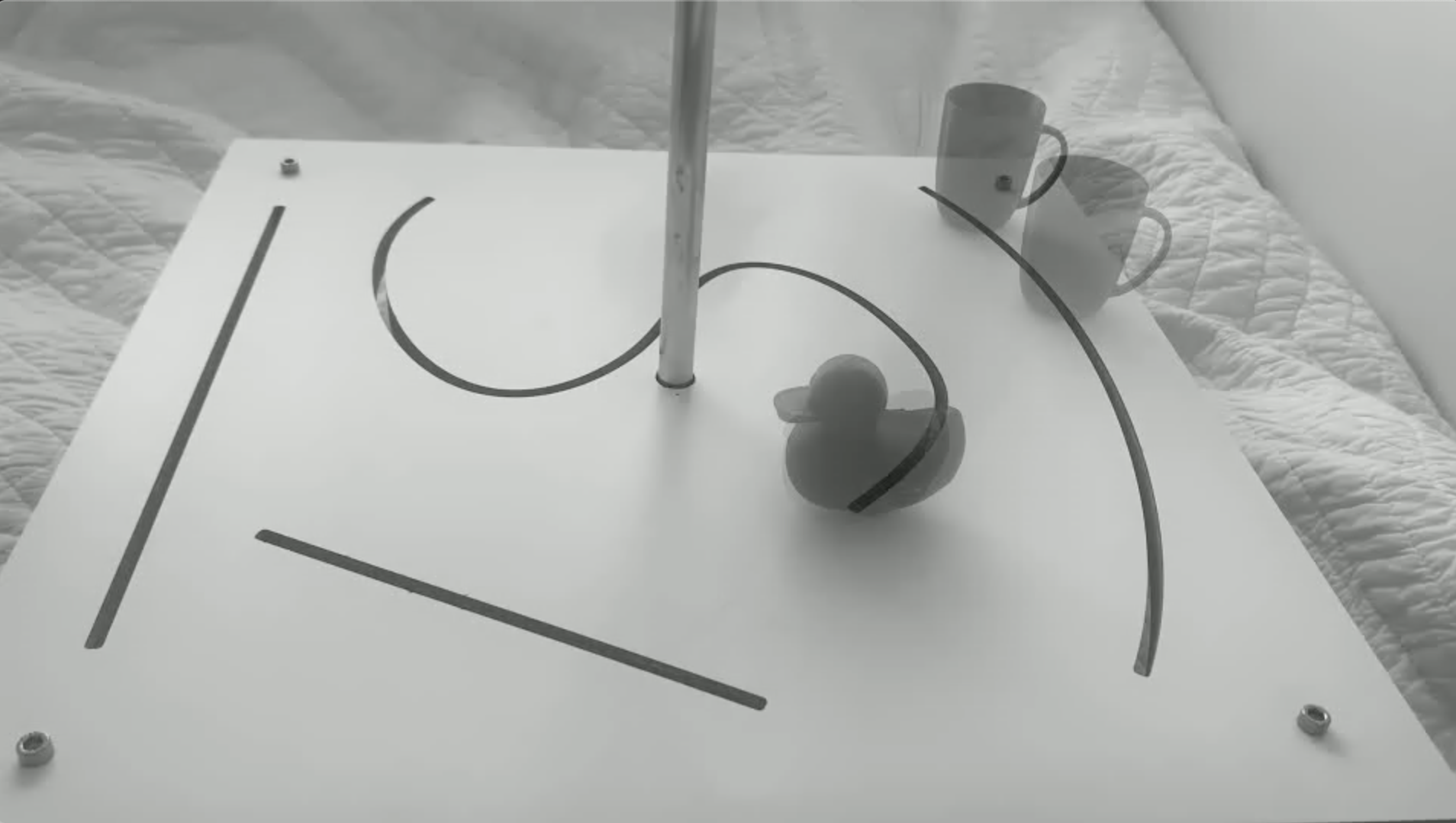}
        \\[\smallskipamount]
        \includegraphics[width=.32\linewidth]{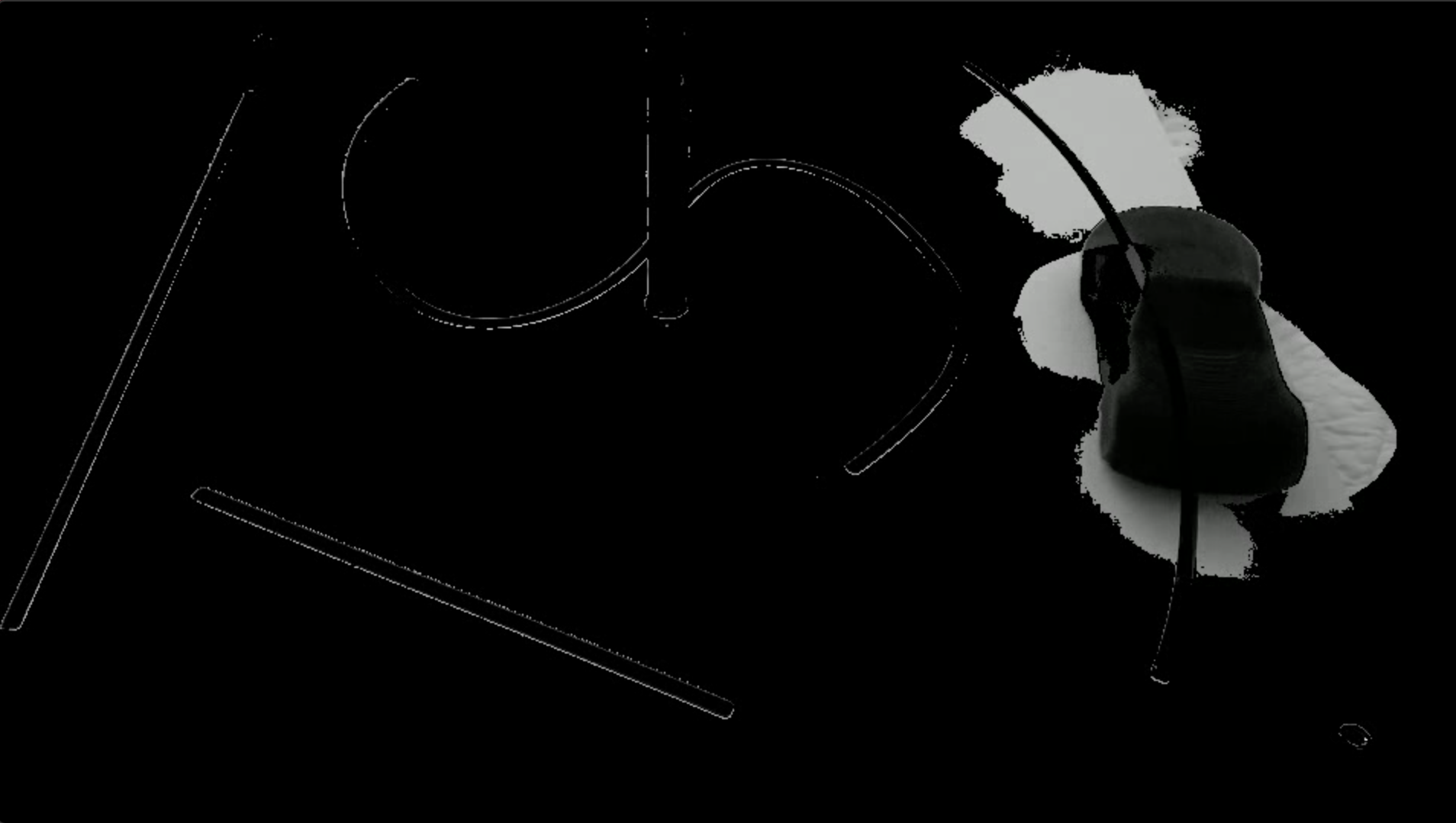}\hfill
        \includegraphics[width=.32\linewidth]{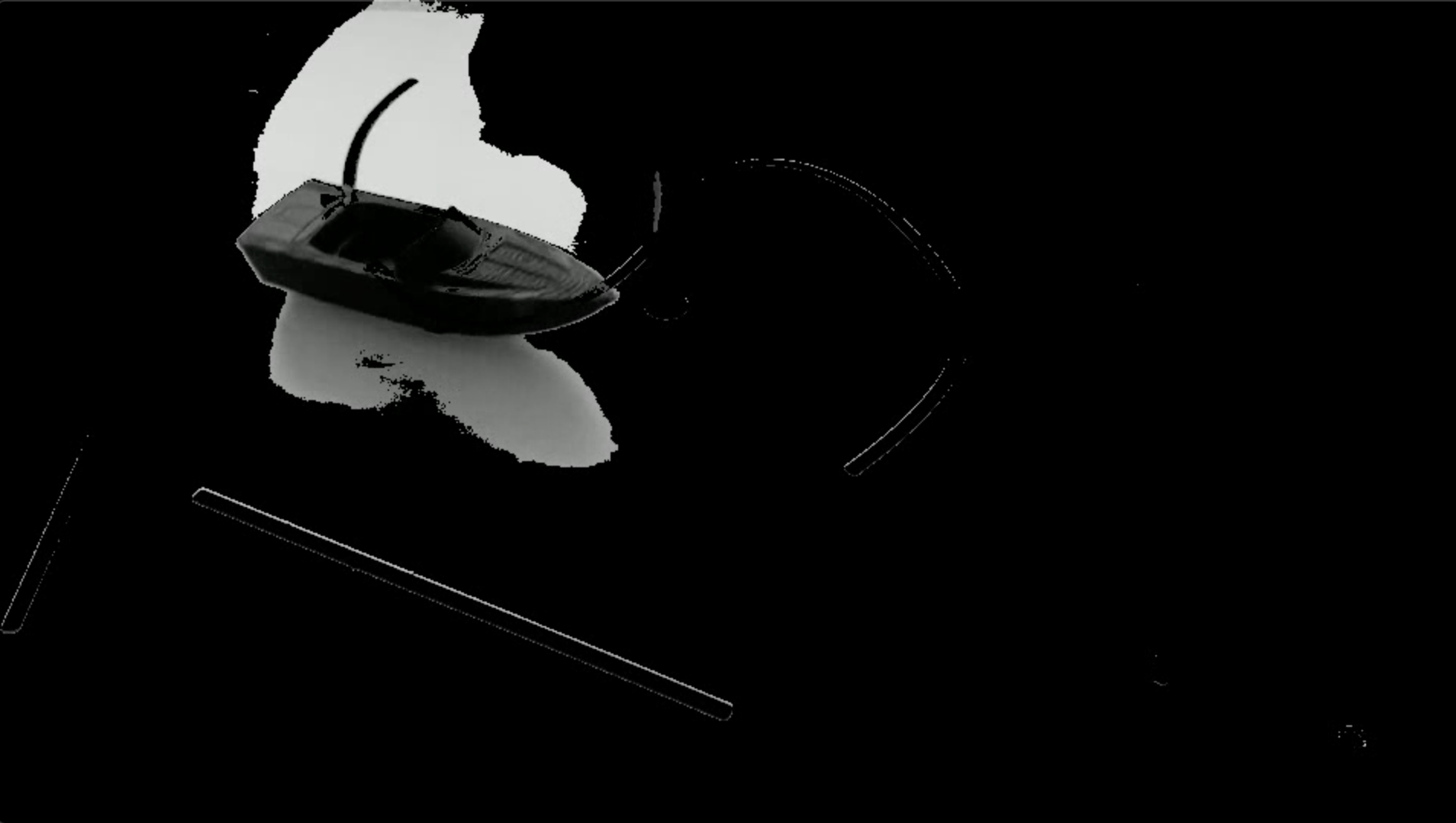}\hfill
        \includegraphics[width=.32\linewidth]{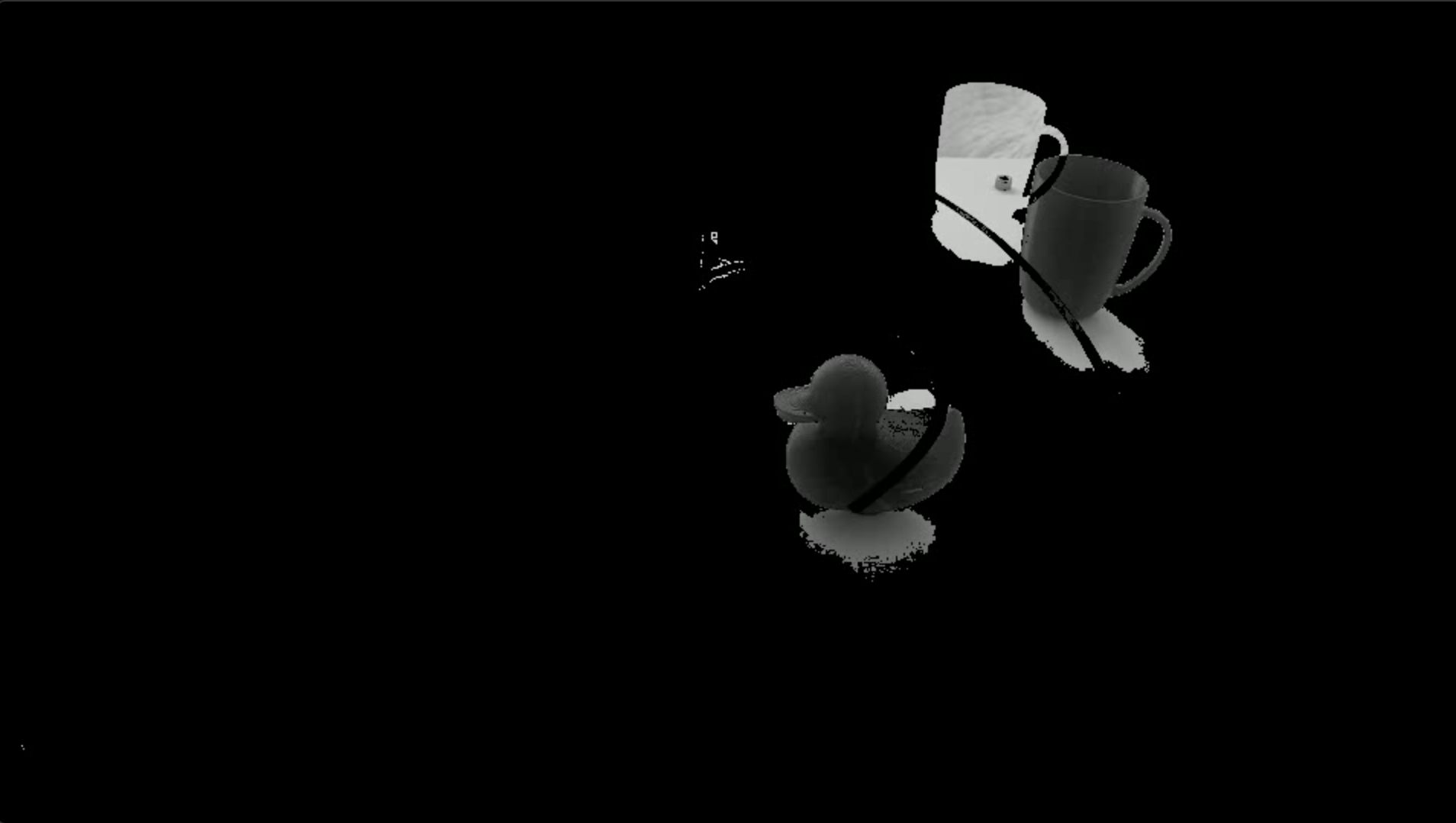}
        \\[\smallskipamount]
        \includegraphics[width=.32\linewidth]{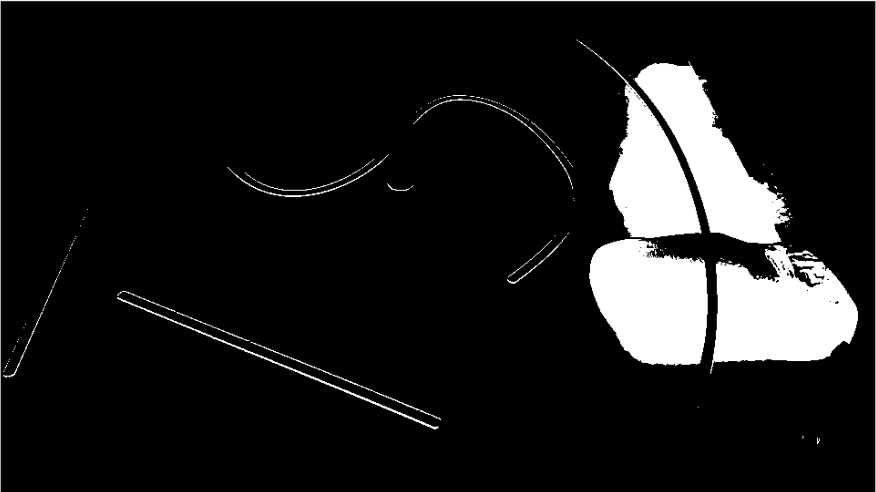}\hfill
        \includegraphics[width=.32\linewidth]{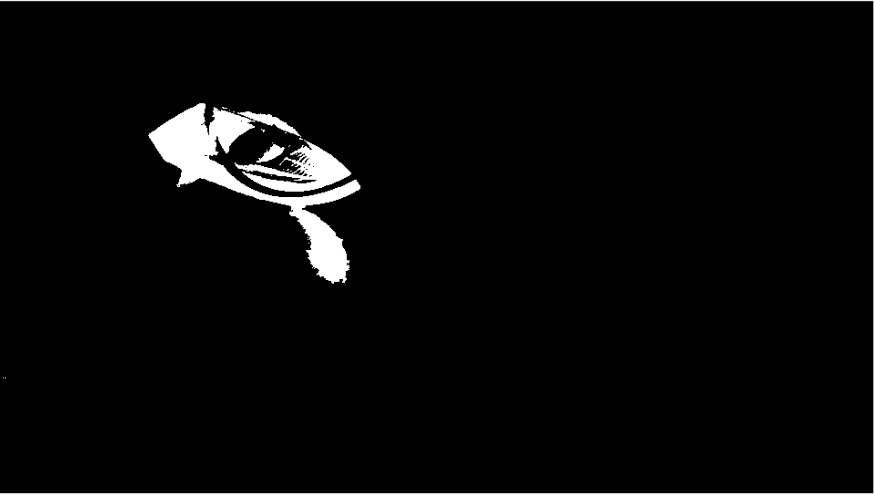}\hfill
        \includegraphics[width=.32\linewidth]{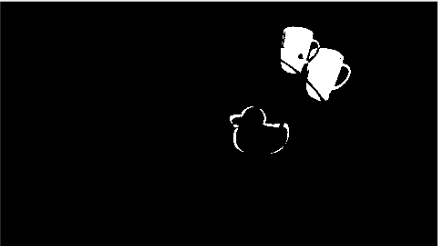}
        \caption{Visual evaluation results for frames from three baseline videos. The top row shows the original baseline video frames. The second and third row shows the predicted static background and the predicted foreground frame, respectively. The fourth row shows the filtered foreground after subtracting the background from the original frame.}
        \label{fig:dmd_results}
    \end{figure}
    
    The video frames were converted to gray-scale and re-scaled by a factor of 0.25 to reduce the memory and computation requirements. Furthermore, the video frames were flattened and stacked as the column vectors of the matrix $\mathbf{X}$ on which the DMD algorithm (as described in \autoref{subsec:method_dmd}) was applied to compute the DMD modes. The computed DMD mode with the lowest oscillation frequency, i.e. smallest changes in time, corresponded to the static background. The static background and the moving foreground (corresponding to moving objects) were separated from each other following the steps earlier explained. The results of background subtraction for three baseline videos from our dataset are presented in  \autoref{fig:dmd_results}. The top row shows an example frame from the original video. In the second row, the predicted background frame is based on the computed DMD modes with lowest oscillation frequency. The third row displays the predicted foreground frame based on the computed DMD modes with high oscillation frequencies. Finally, the fourth row displays the filtered results after extracting the predicted background frame from the original video frame. Looking at the results in the \autoref{fig:dmd_results}, the DMD algorithm appears to successfully classify DMD modes based on their associated high or low frequencies. Judging by the results in the fourth row it can be said that for the baseline videos, the DMD could correctly identify the pixels corresponding to motion.  
    
\begin{figure}
        \begin{subfigure}{\linewidth}
            \includegraphics[width=.49\linewidth]{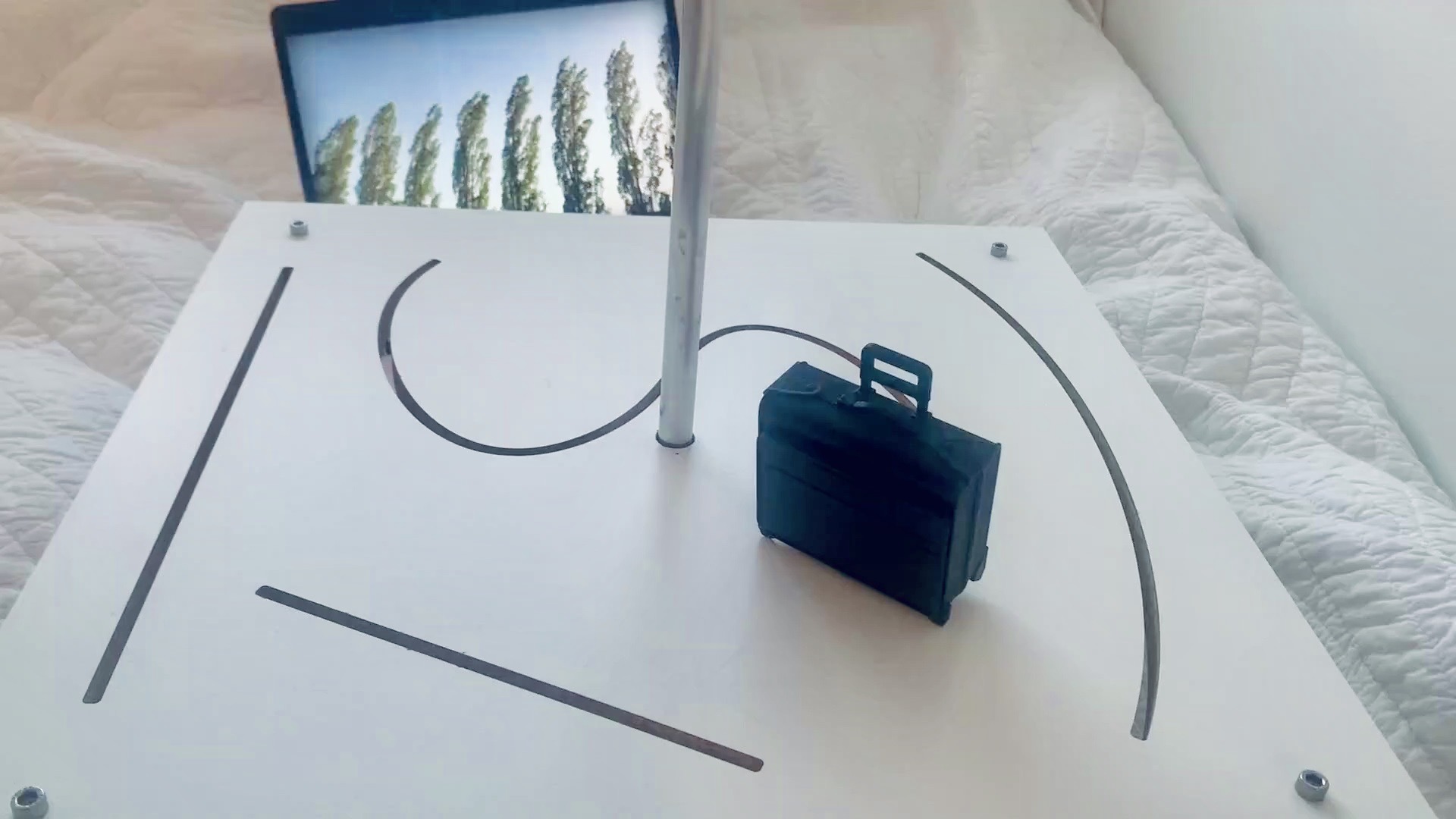}
            \hfill
            \includegraphics[width=.49\linewidth]{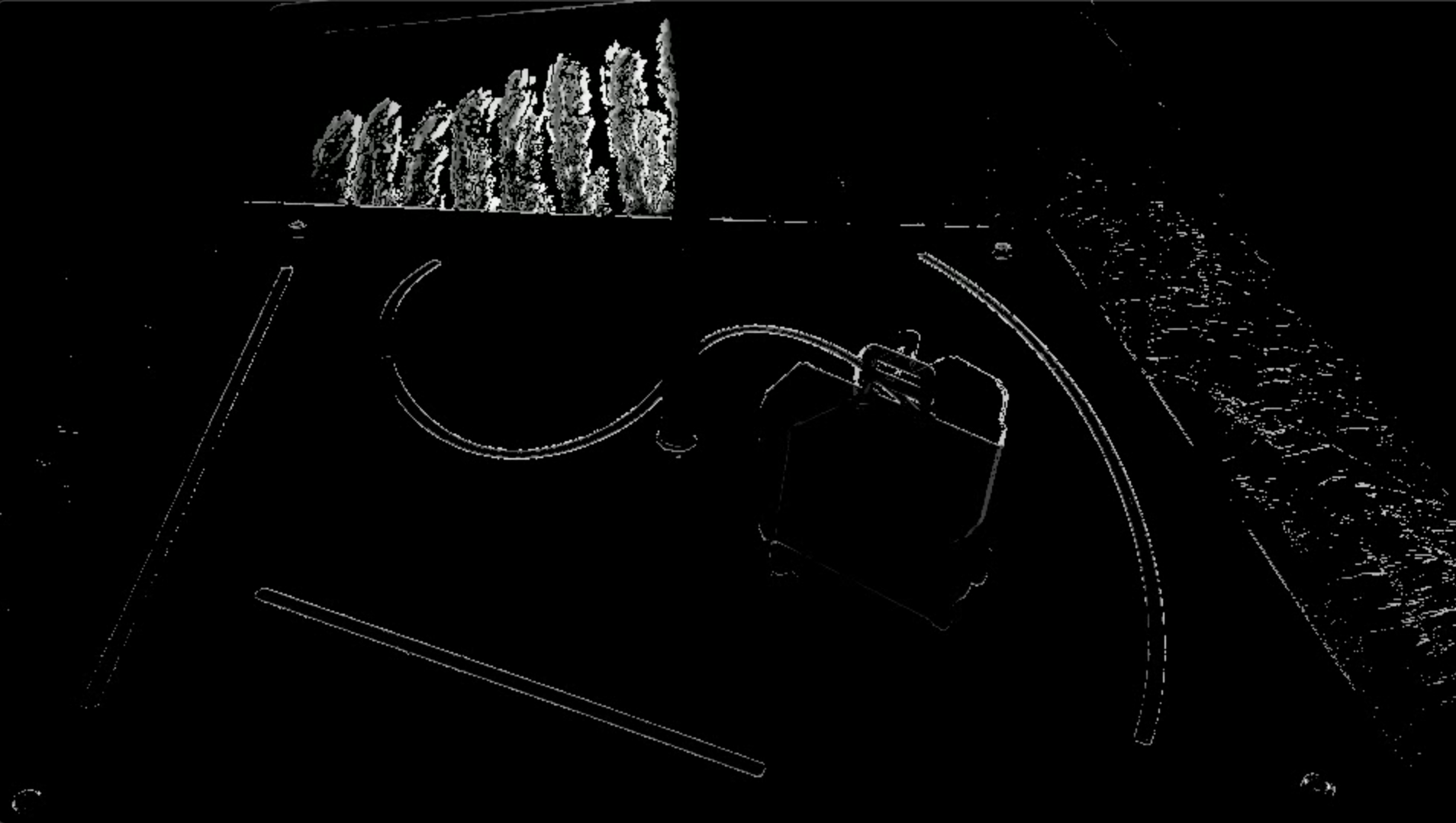}
            \caption{Original video frame to the left and the DMD foreground prediction of the dynamic background to the right.}
            \label{fig:dmd_noise1}
        \end{subfigure}
        \newline
        \newline
        \newline
        \begin{subfigure}{\linewidth}
            \includegraphics[width=.49\linewidth]{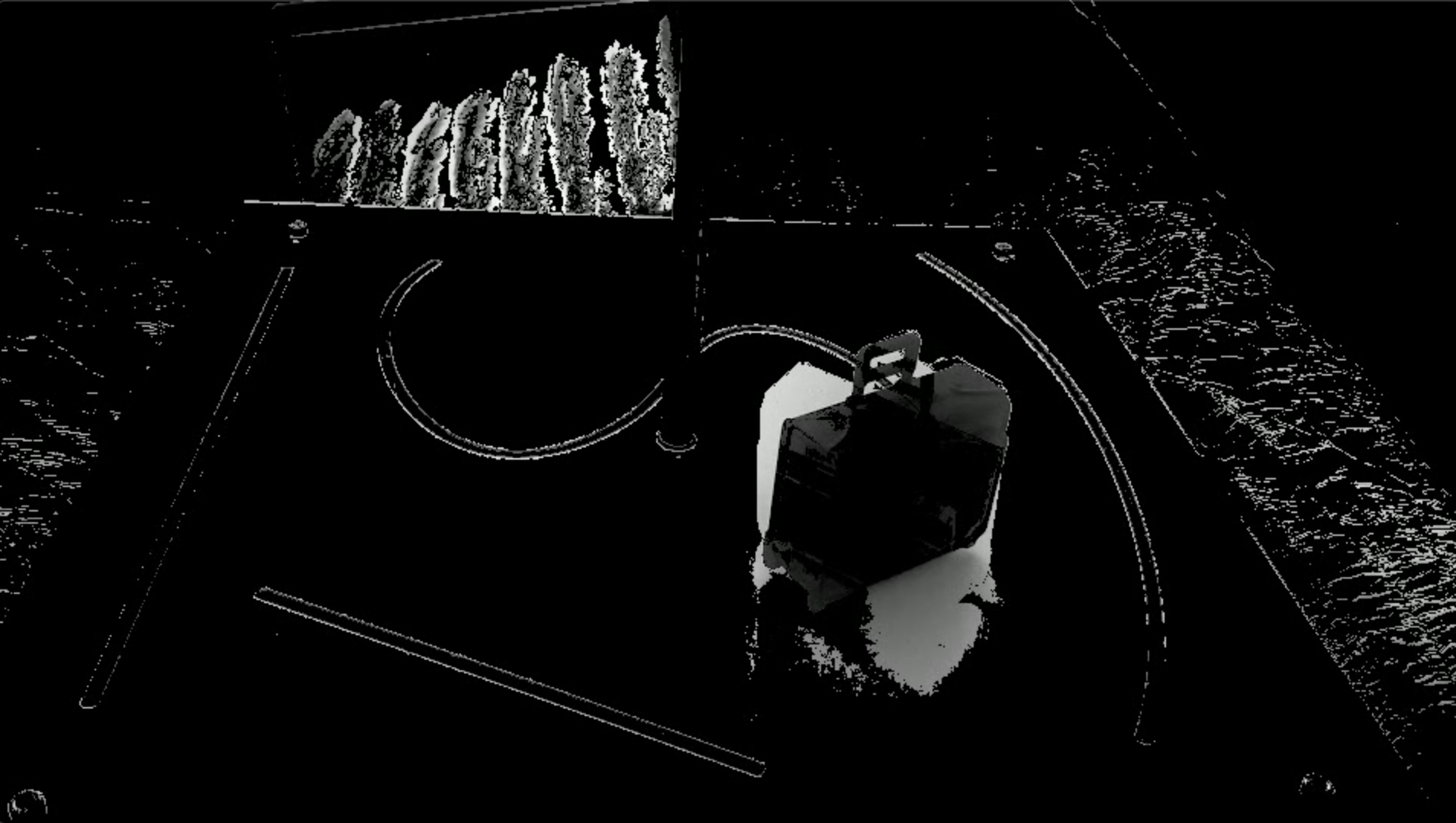}
            \hfill
            \includegraphics[width=.49\linewidth]{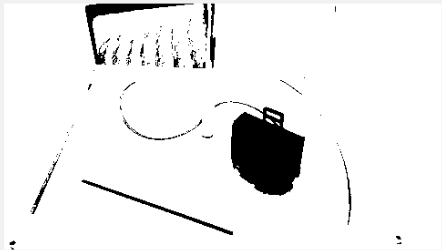}
            \caption{DMD foreground prediction of the dynamic background and the moving 3D object to the left, and the filtered foreground to the right.}
            \label{fig:dmd_noise2}
        \end{subfigure}
        \caption{DMD results for a scene subjected to noise from a dynamic background}
        \label{fig:dmd_noises}
    \end{figure}

    Next the DMD algorithm was applied to the videos recorded under subtle background movements (a screen displaying moving trees). The results are presented in \autoref{fig:dmd_noises}. It can be clearly seen that the disturbances in the background of the scene can also be detected as foreground motion. Fortunately, this situation should not arise in the context of digital twin as everything in the scene will be considered a part of it and all such motions will be relevant. 
    
\begin{figure}
        \includegraphics[width=.49\linewidth]{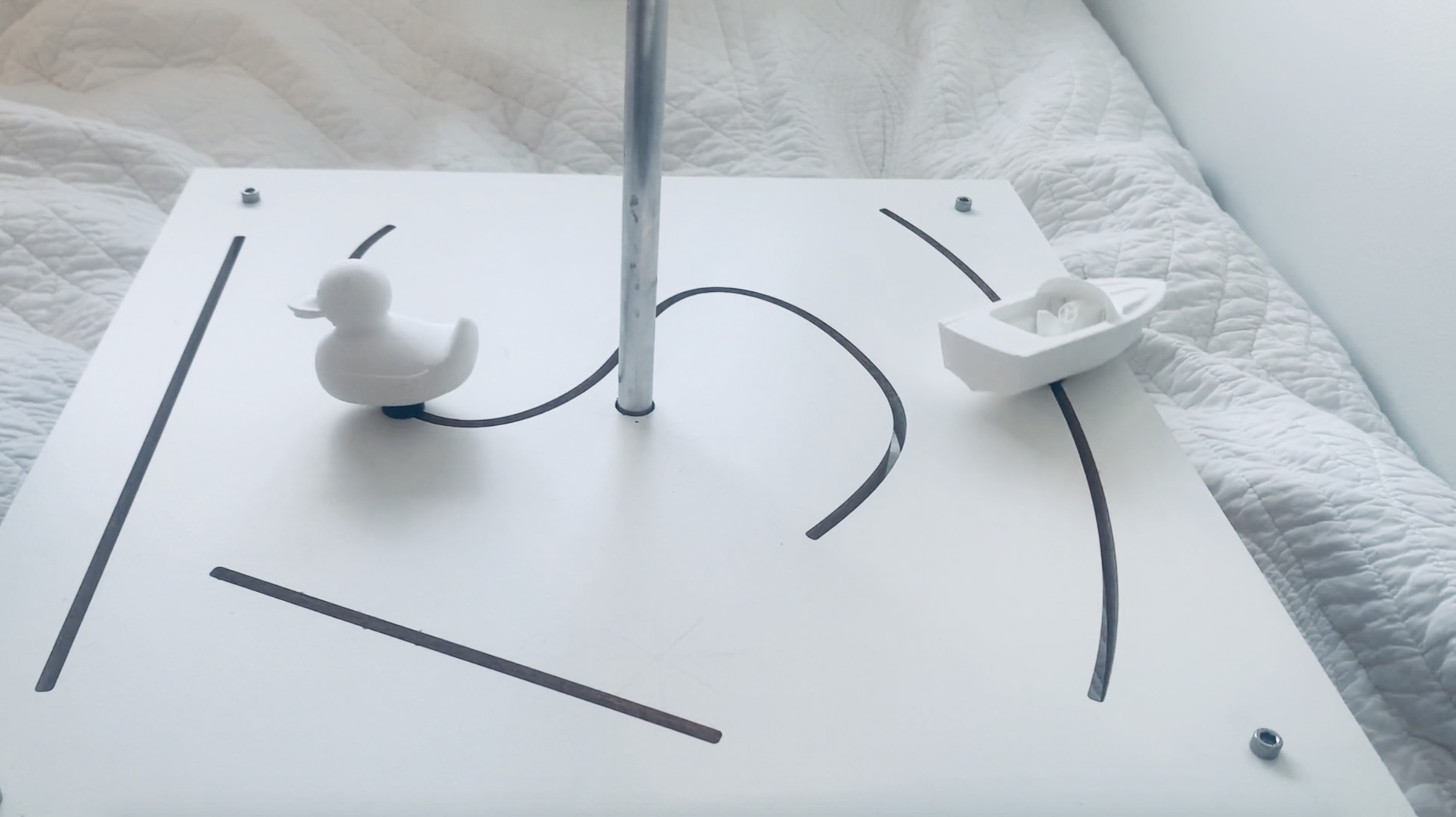}\hfill
        \includegraphics[width=.49\linewidth]{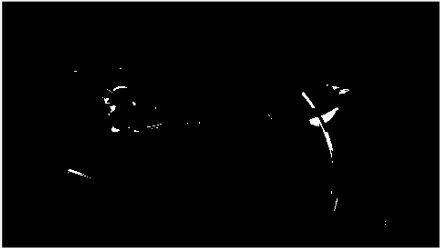}
        \caption{DMD results on foreground object with same pixel intensity as background object. The original frame is displayed to the left and the filtered foreground is displayed to the right.}
        \label{fig:dmd_white}
    \end{figure}

\begin{figure}
        \includegraphics[width=.49\linewidth]{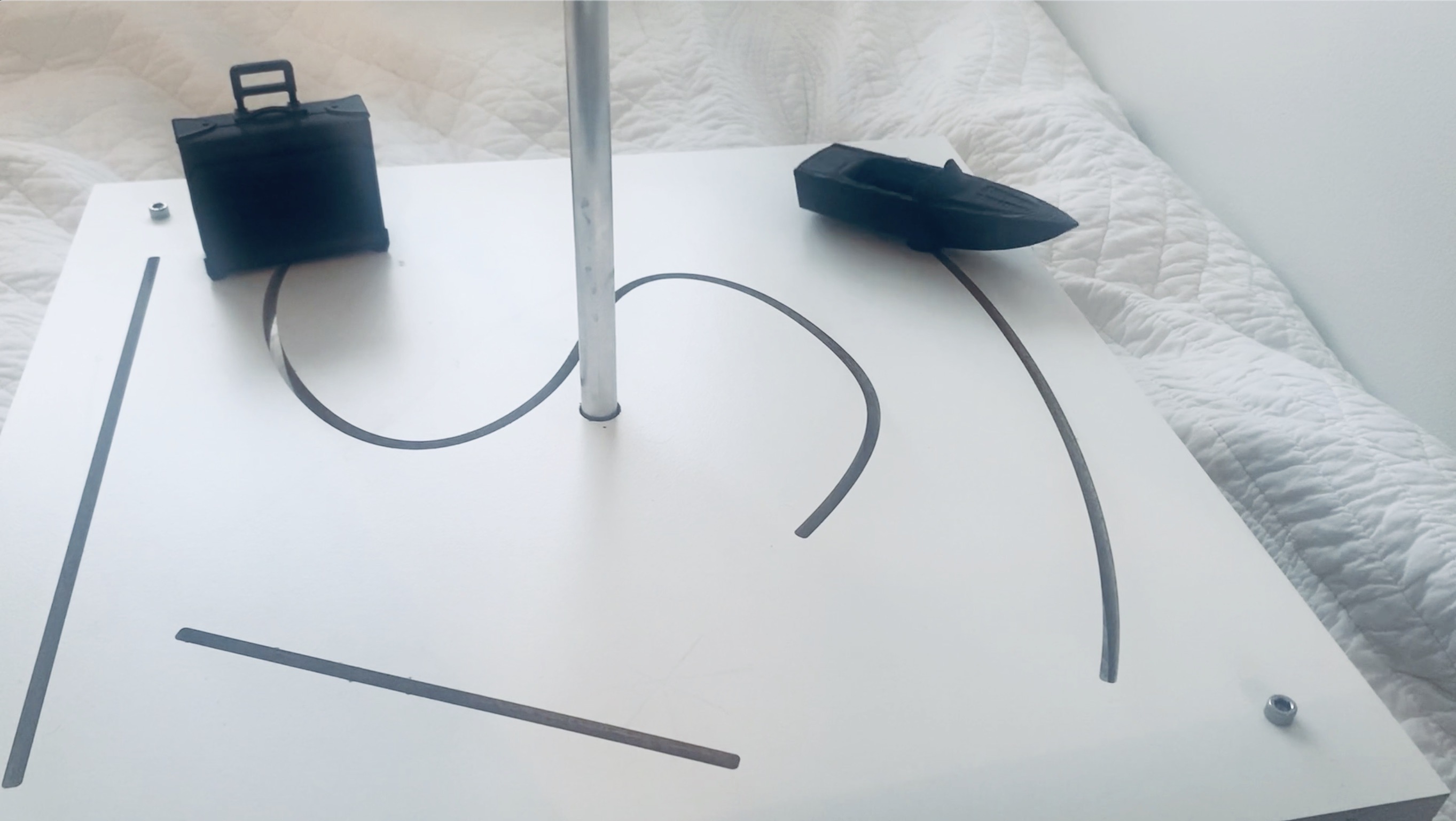}\hfill
        \includegraphics[width=.49\linewidth]{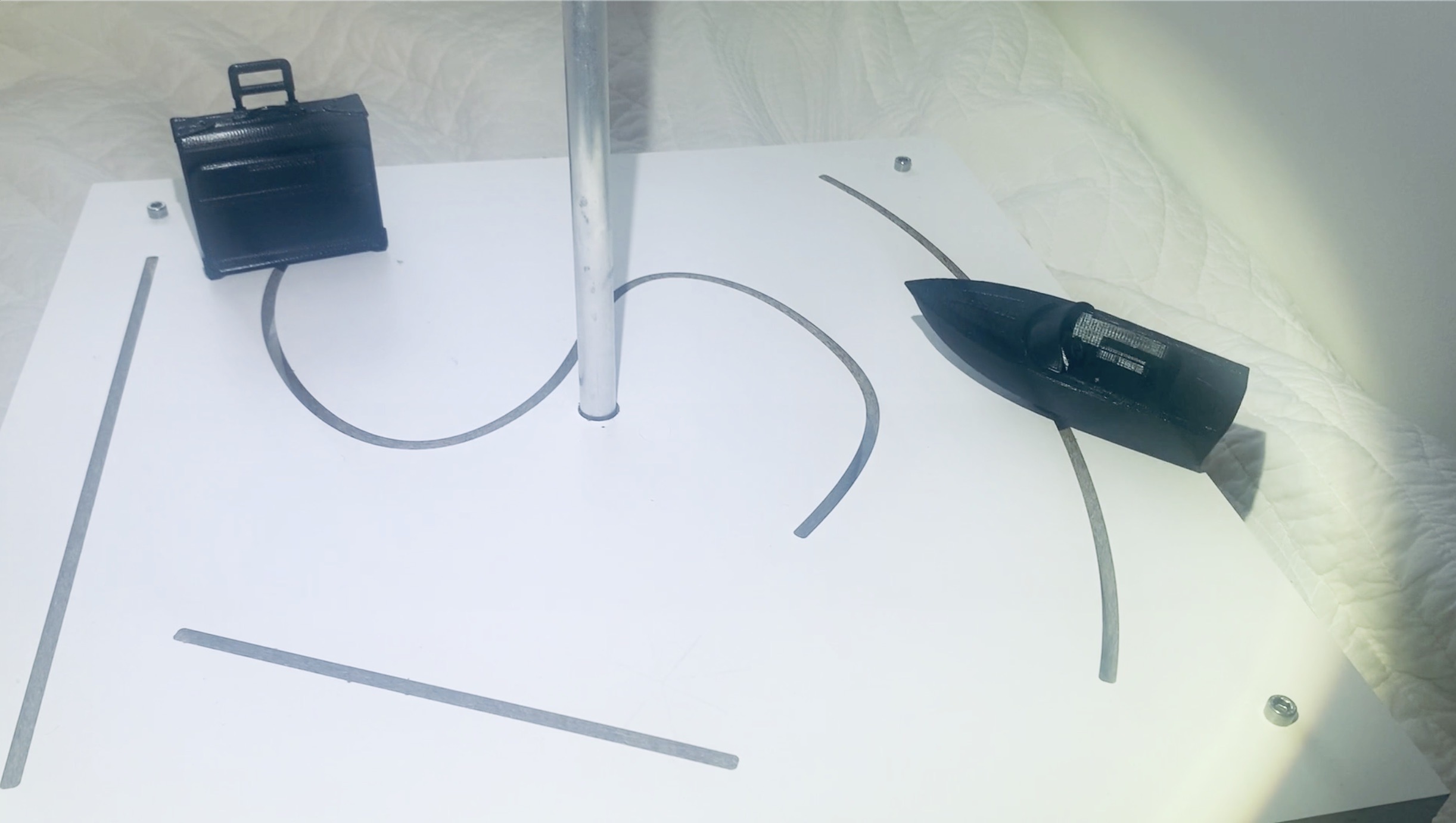}
        \\[\smallskipamount]
        \includegraphics[width=.49\linewidth]{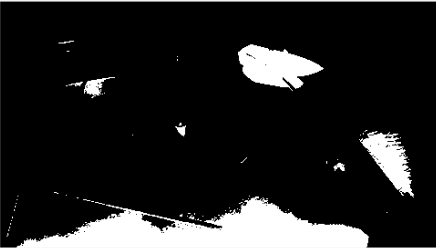}\hfill
        \includegraphics[width=.49\linewidth]{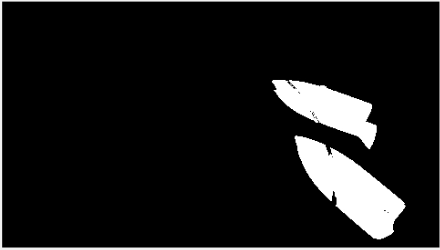}
        \caption{DMD results for a scene subjected to sudden changes in lighting conditions. The first row displays original frames, and the second row displays filtered foreground frames.}
        \label{fig:dmd_light_change}
    \end{figure}

    Another challenge for DMD based motion-detection is the method's ability to detect moving foreground objects with the same pixel intensity as the background, i.e. monochrome frames. The results corresponding to this situation are presented in \autoref{fig:dmd_white}. Here, we see that it has been difficult to identify the foreground objects (duck and boat), when the pixel intensity of foreground objects is similar to background. Furthermore, \autoref{fig:dmd_light_change} presents the results corresponding to a sudden change in the lighting conditions. The changes in lighting condition resulted in detections quite similar to that observed in the baseline results. However, such sudden changes in the lighting condition is a disturbance that can be detected by additional light sensors to detect false motions.
    
    The results presented so far displays some of the strengths and limitations of the DMD algorithm for detecting motion in the realistic data-set obtained from our experimental set-up. In the context of digital twins, it has been able to detect motion well for baseline video frames, and the known challenging scenarios can be mitigated to an extent by proper planning in the operational environment like using a light sensor or making the full scene a part of the DT.
    
    Next, we use object detection on the last frame of the interval over which motion was detected by the DMD algorithm. 
    
    \subsection{Object Detection using YOLOv5}
    \label{subsec:yolo}
    The object detection algorithm YOLOv5 was implemented in PyTorch and retrained for 300 epochs on the dataset of 463 images (and their augmented versions) collected from our experimental set-up. The model outputs images containing labeled bounding boxes for all the detected objects in an image. The performance of the the retrained network was measured by comparing against the ground truth, i.e., the manually labeled images. A precision-recall curve and training loss curves were used as diagnostic tools for deciding when the model was sufficiently trained.
    \begin{figure}
    \centering
        \includegraphics[width=0.8\linewidth]{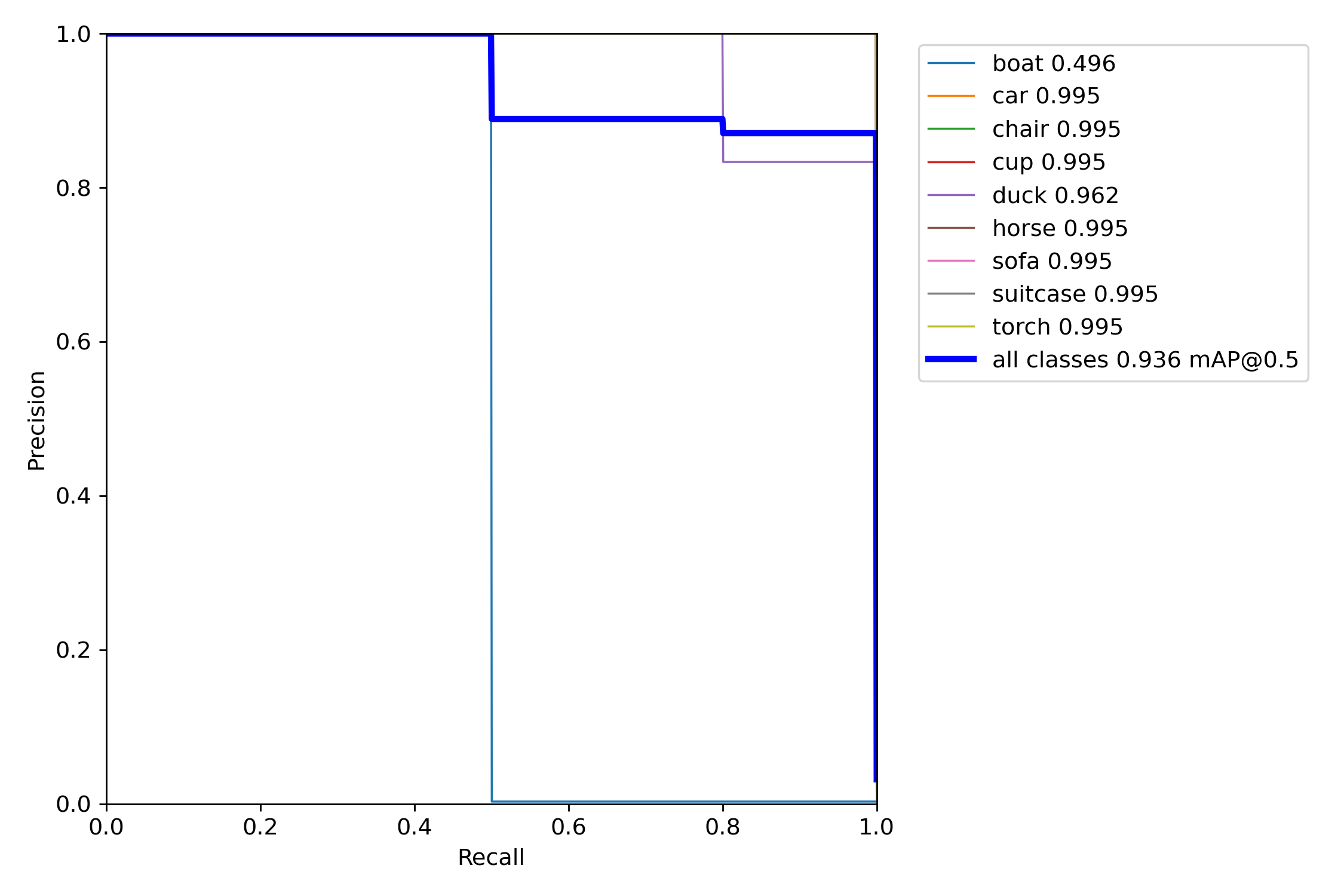}
        \caption{Precision-recall curve for our nine object categories from YOLOv5}
        \label{fig:precision_recall_curve}
    \end{figure}
    \autoref{fig:precision_recall_curve} displays the plotted precision-recall curve for our object detection network on the chosen object categories. The overall mean network performance on all object classes resulted in a high mAP of 0.936. This is a good result clearly marked by the precision-recall curve plotted as a thick blue line. Most object categories barring the boat category obtained high mAP results, resulting in a high average score.   
    \begin{figure}
        \includegraphics[width=0.32\linewidth]{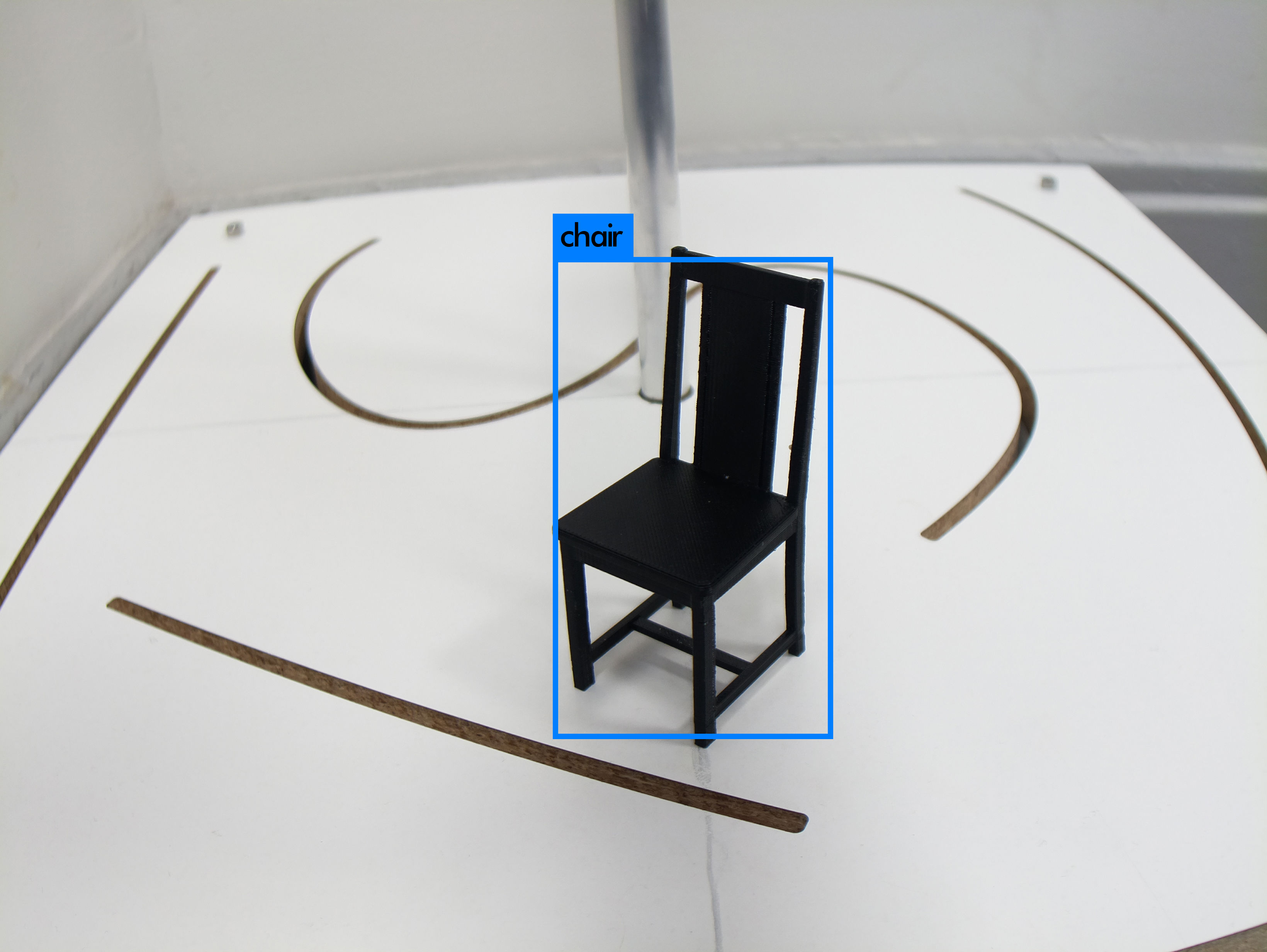}
        \hfill
        \includegraphics[width=0.32\linewidth]{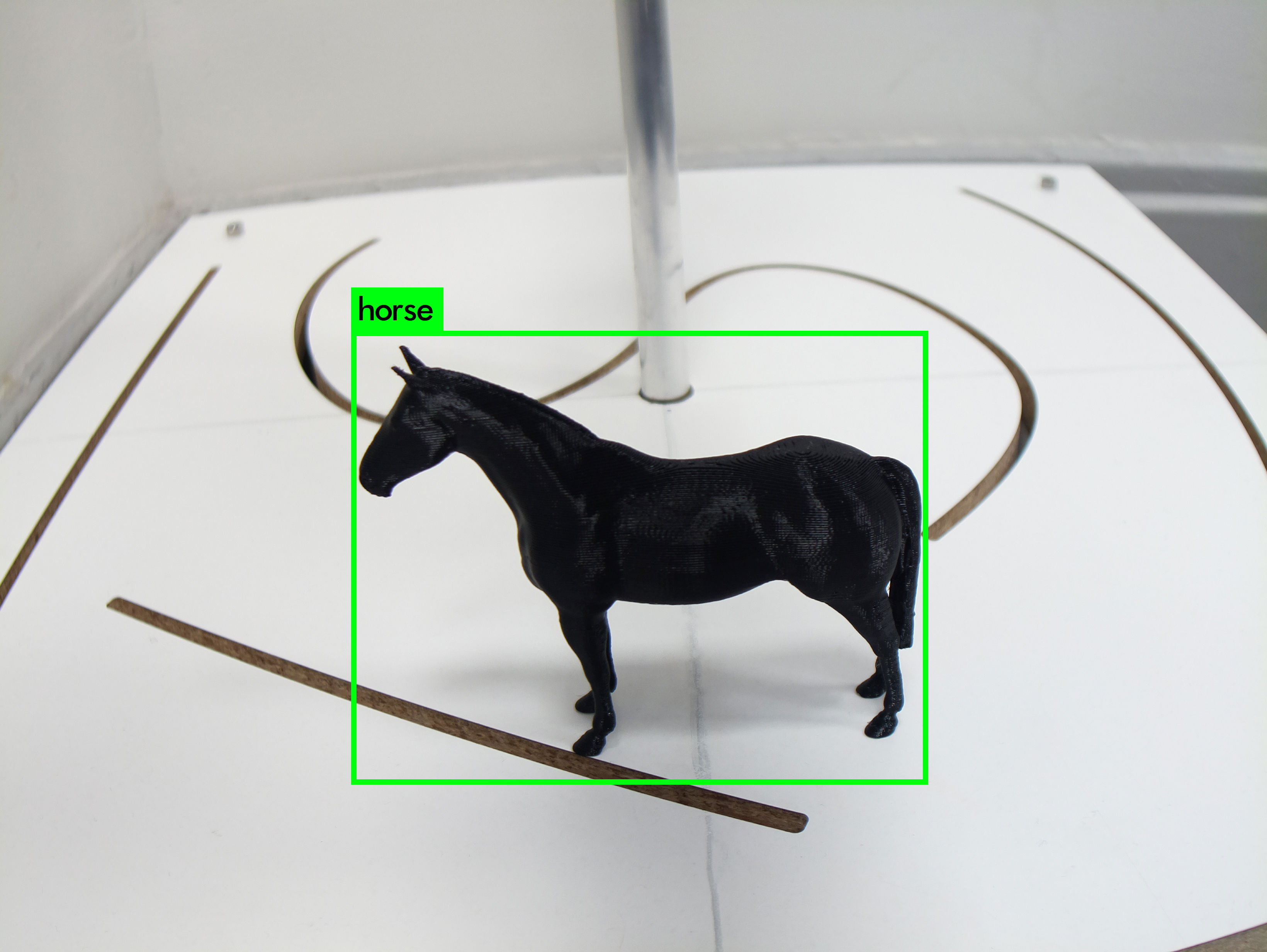}
        \hfill
        \includegraphics[width=0.32\linewidth]{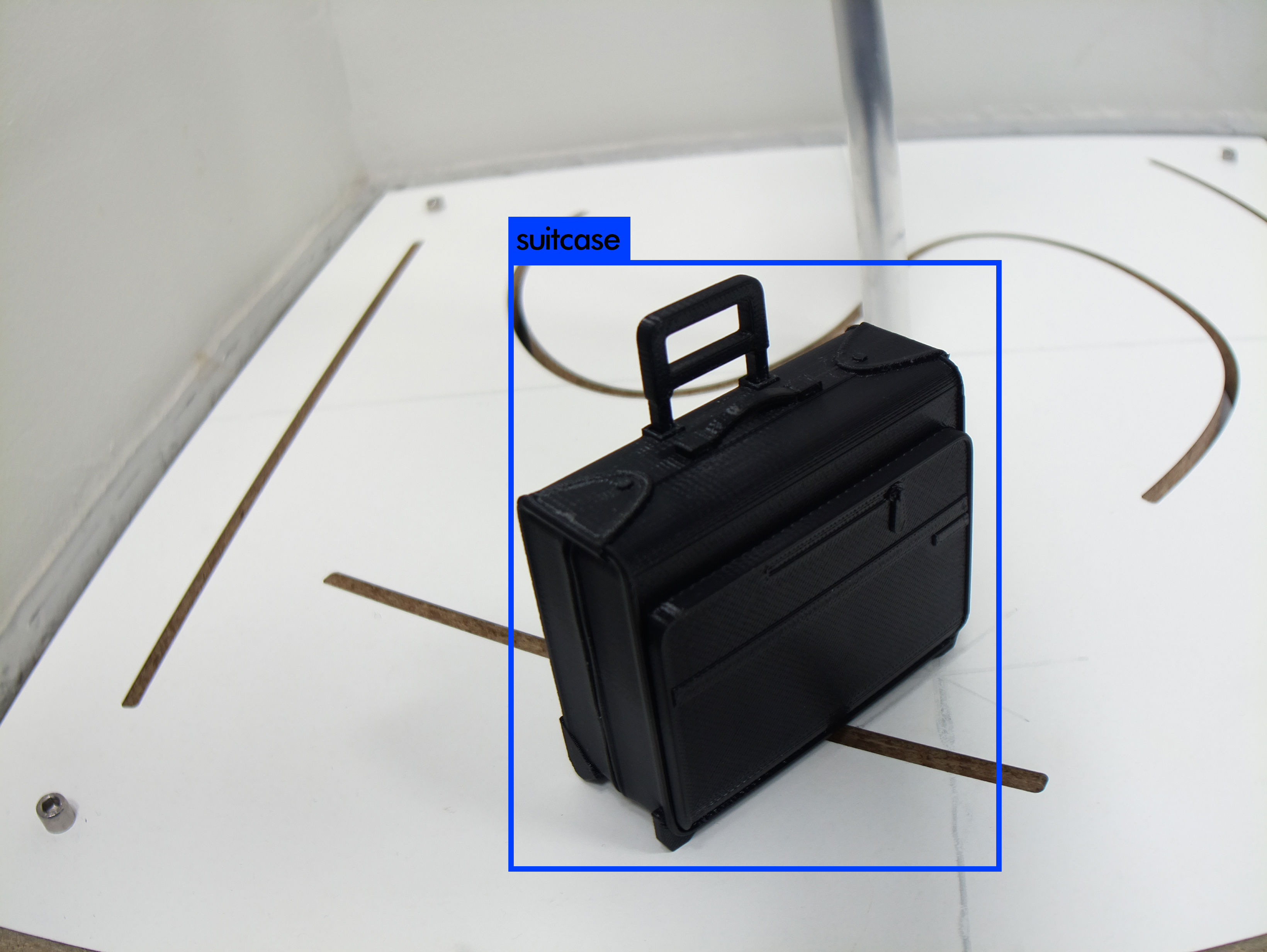}
        \caption{Examples of objects detected by YOLOv5}
        \label{fig:yolo_bounding_box_examples}
    \end{figure}
    
    The \autoref{fig:yolo_bounding_box_examples} illustrates an example of the detected objects and their respective bounding boxes. Furthermore, we know, based on several previously published sources, that the YOLO network can be trained to output close to perfect predictions given proper training data and the right configurations. We would like to stress that in the context of digital twin we might already know most of the objects that can feature in it and hence we can always retrain the object detection algorithm with the new objects to achieve even higher performance. In the current work, the purpose of the object detection and classification is to extract the bounding boxes including the images to be fed to the 3D pose estimation module and YOLO seems to achieve this objective satisfactorily. 
    
    \subsection{3D Pose Estimation}
    \label{subsec:result_poseestimation}
    Following the video frame extraction from our motion detection algorithm, we evaluated the performance of the implemented pose estimation algorithm. The algorithm we used builds upon a network that does not require training on specific categories used for testing \citep{xiao_pose_2019}. The intention is for the estimation method to work on novel object categories. The method requires a 3D CAD model in OBJ format as input, used for rendering images of each object from different angles, as explained in \autoref{subsec:poseestimation}. It also requires the input of a cropped out bounding box containing the object used for pose estimation. This is provided by the trained YOLOv5 network. The results are presented as the percentage error which is defined as 
    \begin{equation}
        \text{Percentage error} = \frac{\mid \Delta \phi_{real}-\Delta \phi_{pred}\mid}{\Delta \phi_{real}}*100
        \label{eq:azi_percentage_error}
    \end{equation}
    \begin{figure}
    \centering
    \includegraphics[width=.12\linewidth]{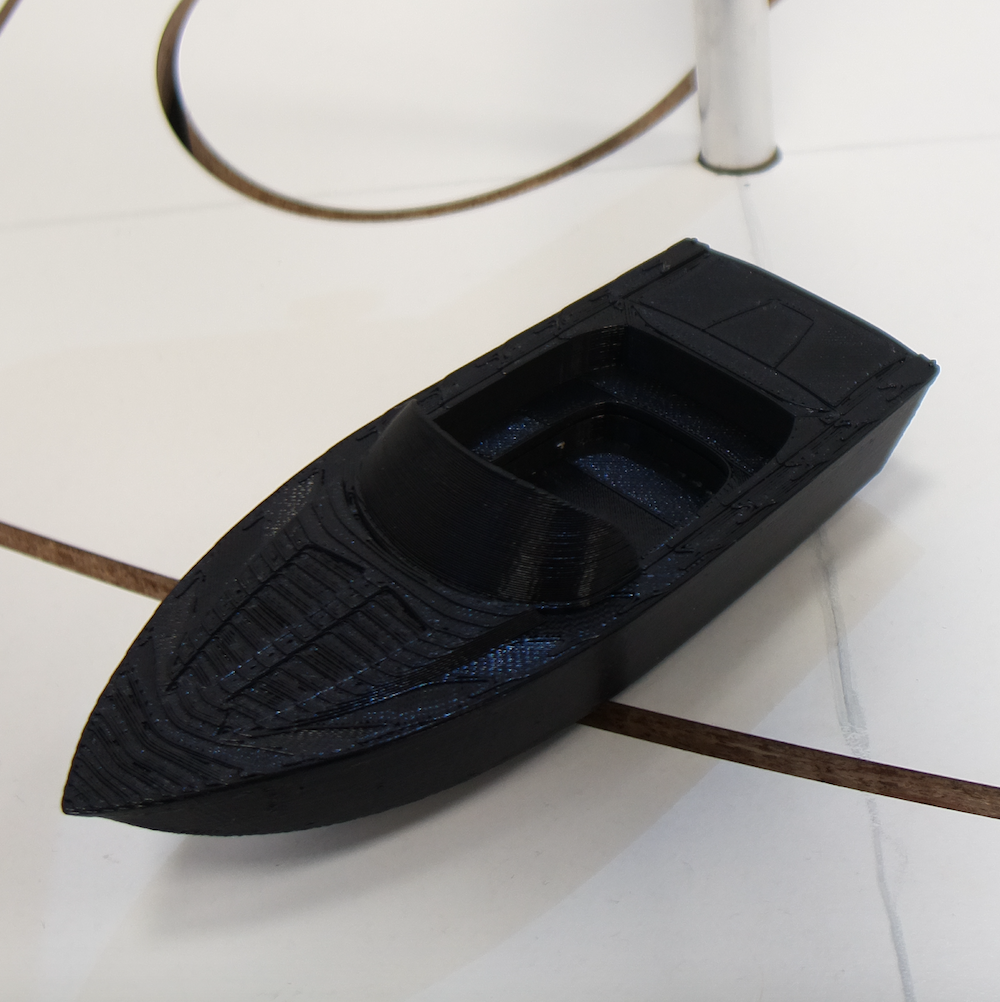}
    \includegraphics[width=.12\linewidth]{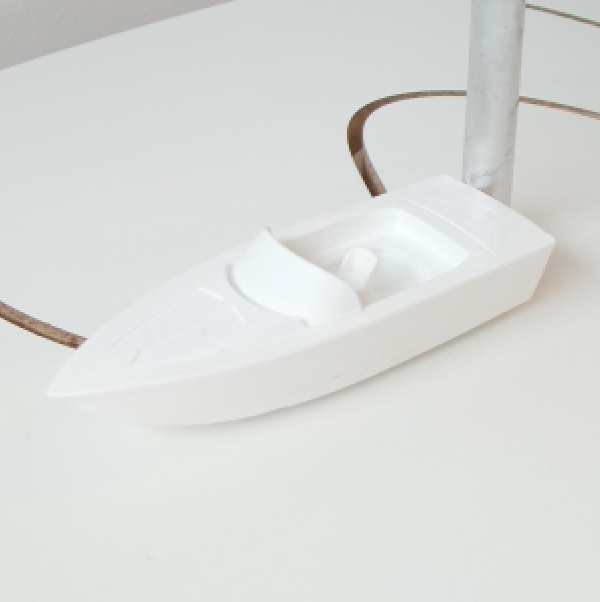}
    \includegraphics[width=.12\linewidth]{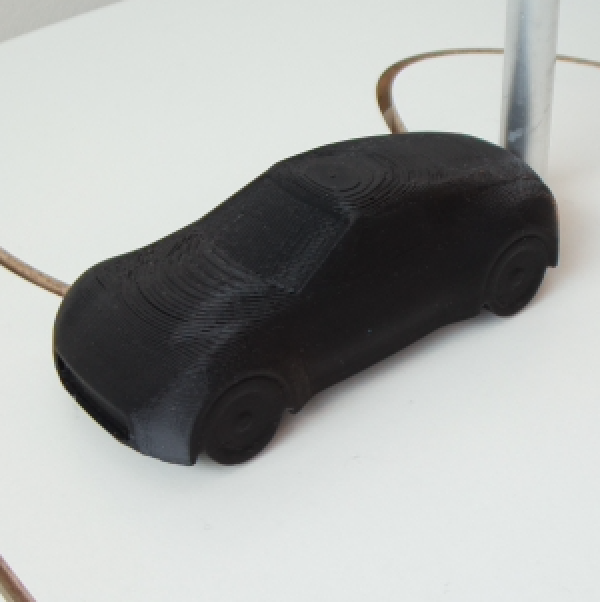}
    \includegraphics[width=.12\linewidth]{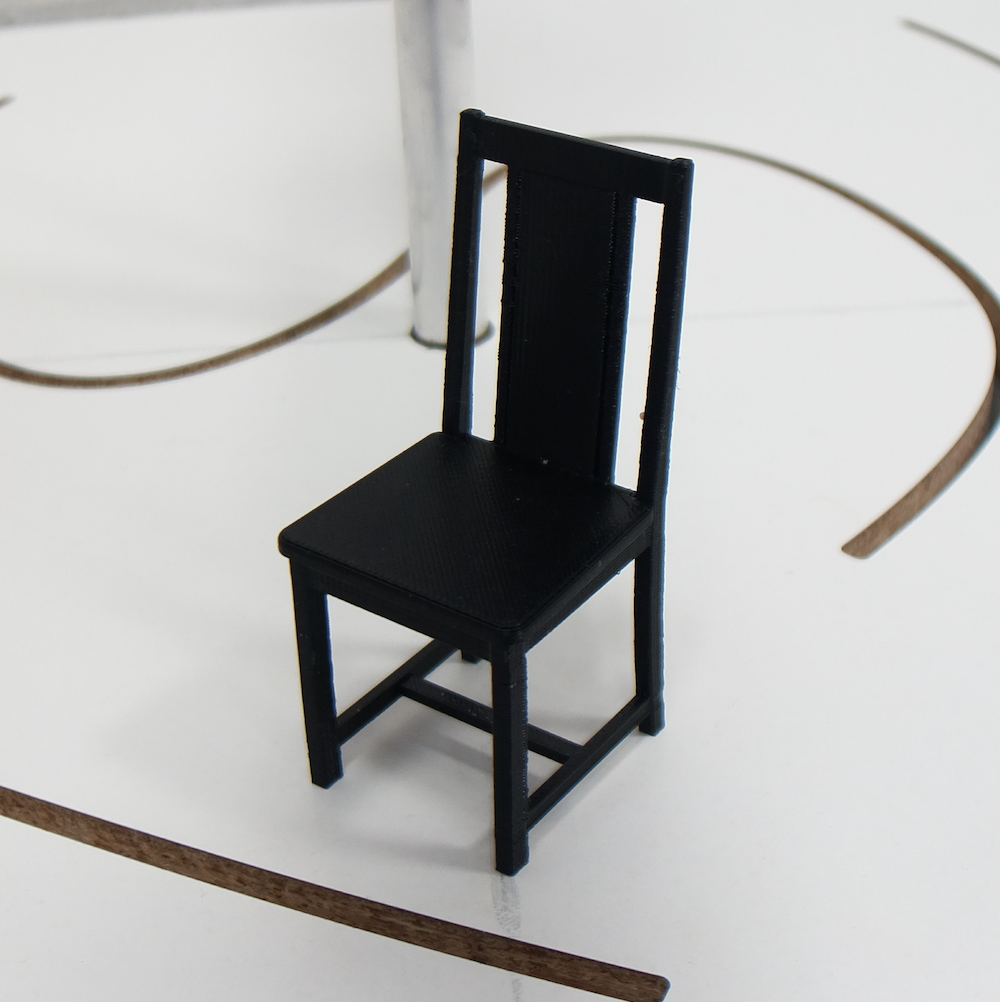}
    \includegraphics[width=.12\linewidth]{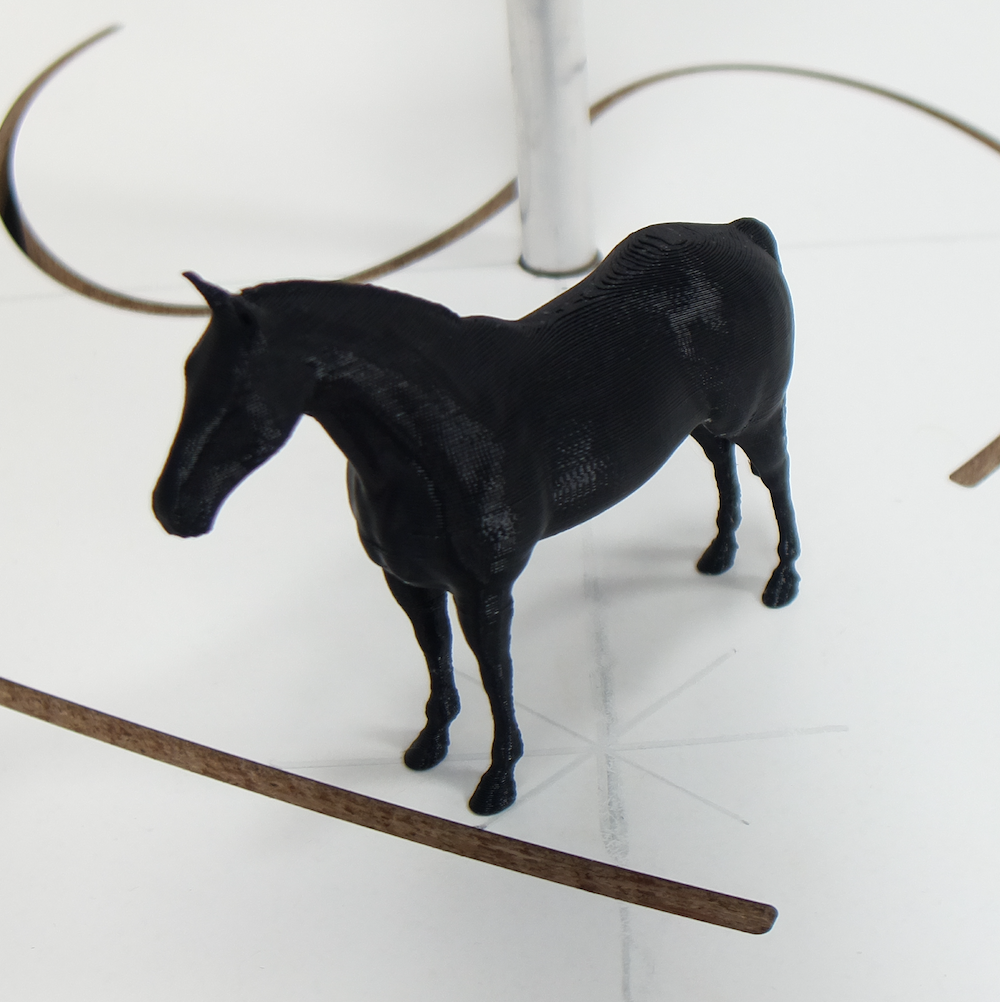}
    \includegraphics[width=.12\linewidth]{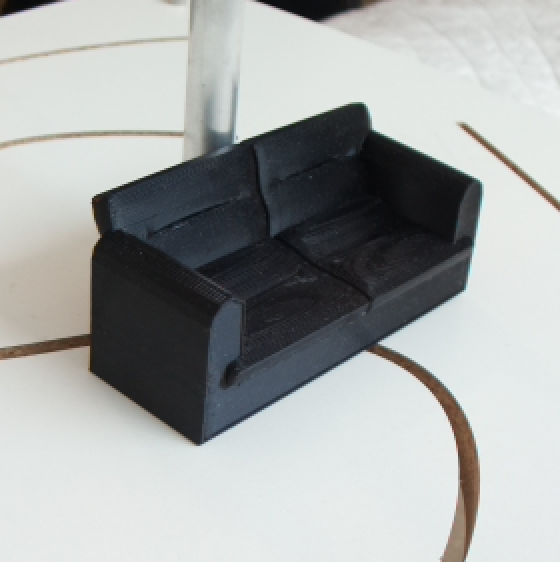}
    \\[\smallskipamount]
    \includegraphics[width=.12\linewidth]{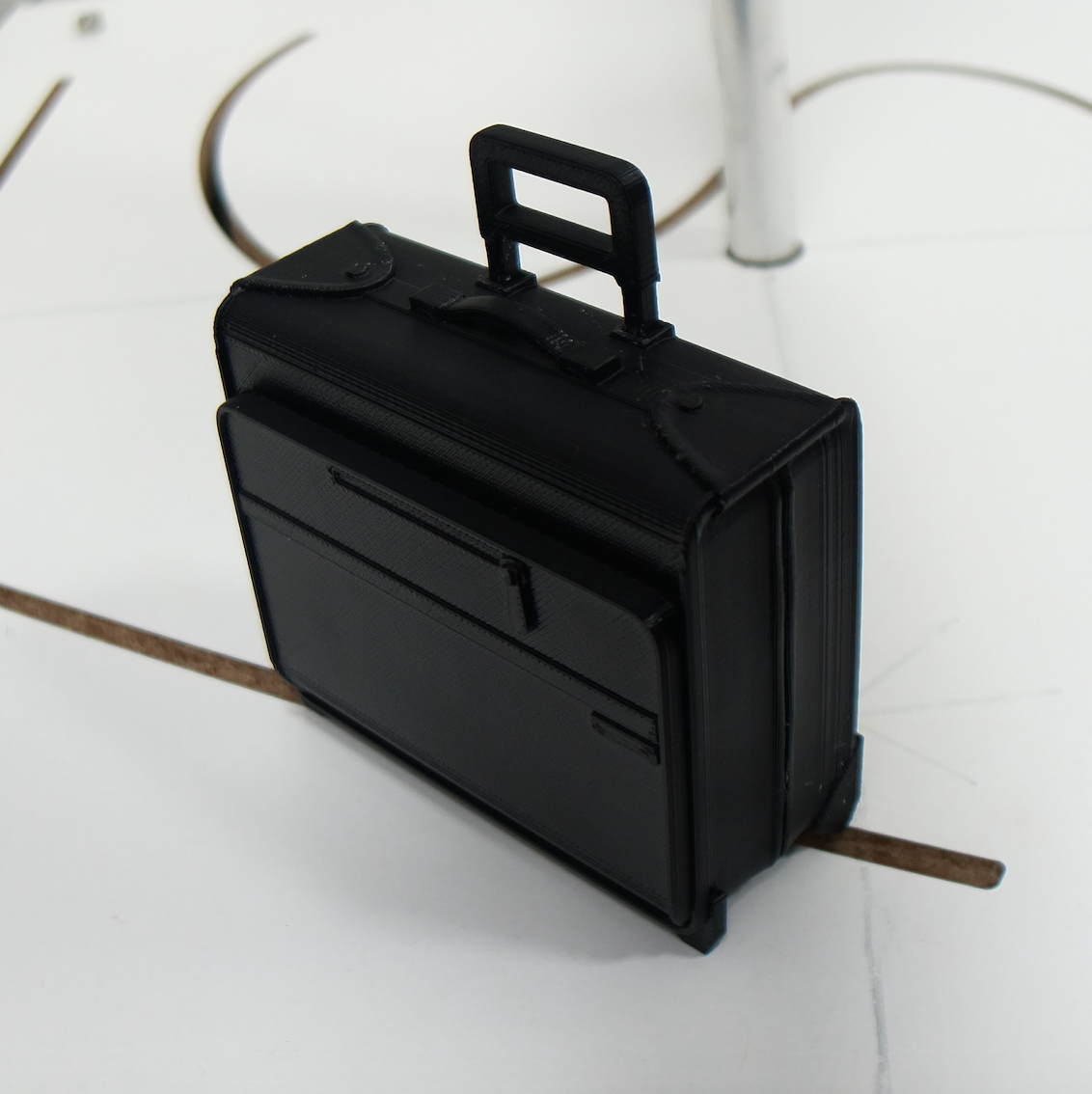}
    \includegraphics[width=.12\linewidth]{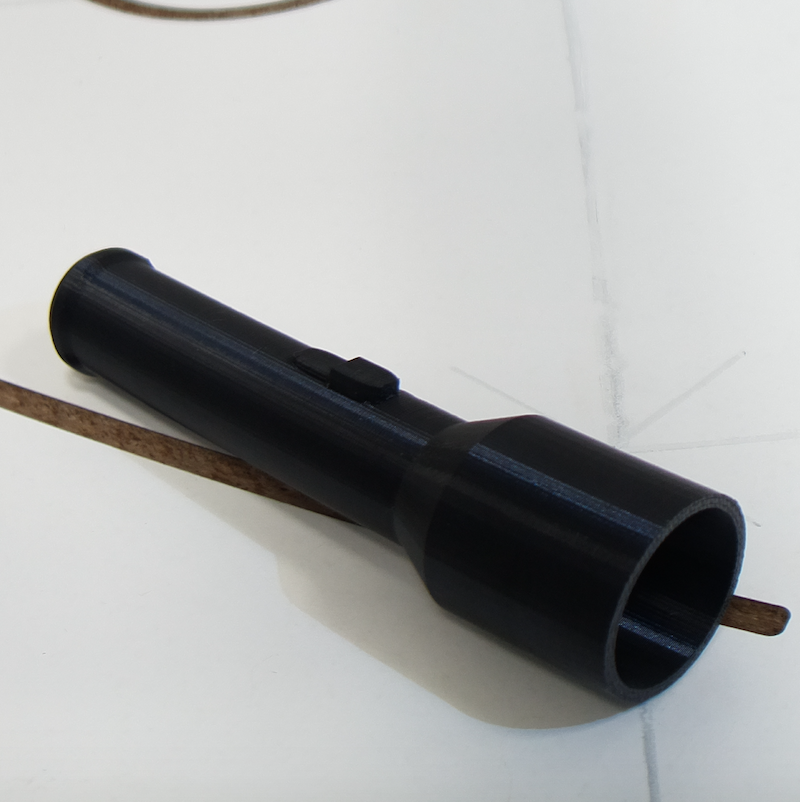}
    \includegraphics[width=.12\linewidth]{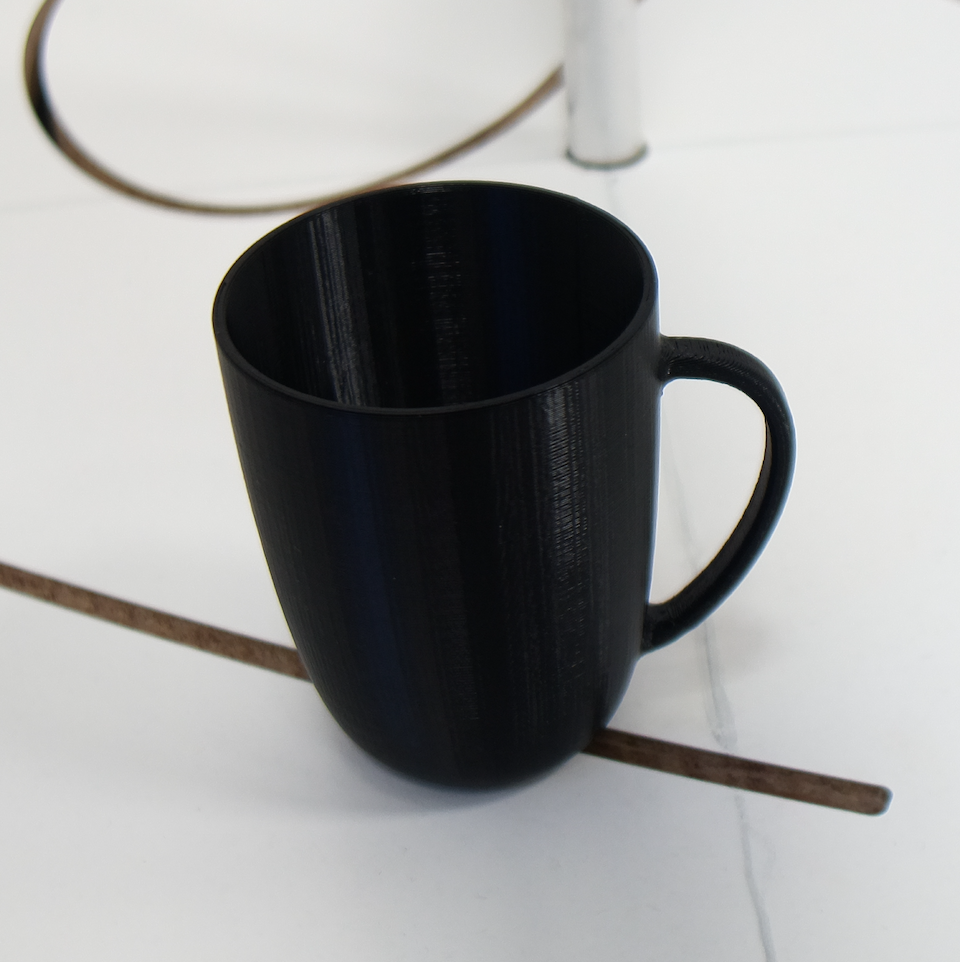}
    \includegraphics[width=.12\linewidth]{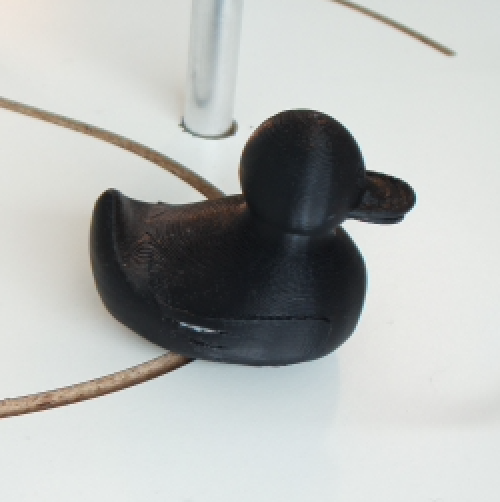}
    \includegraphics[width=.12\linewidth]{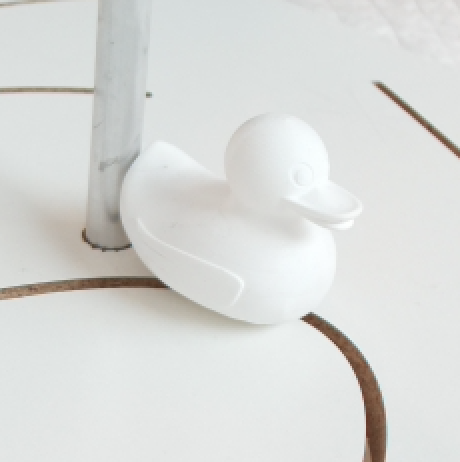}
    \caption{3D printed CAD models used in experiments}
    \label{fig:reference_models}
\end{figure}

\begin{figure}
    \begin{subfigure}[b]{\linewidth}
        \includegraphics[width=\linewidth]{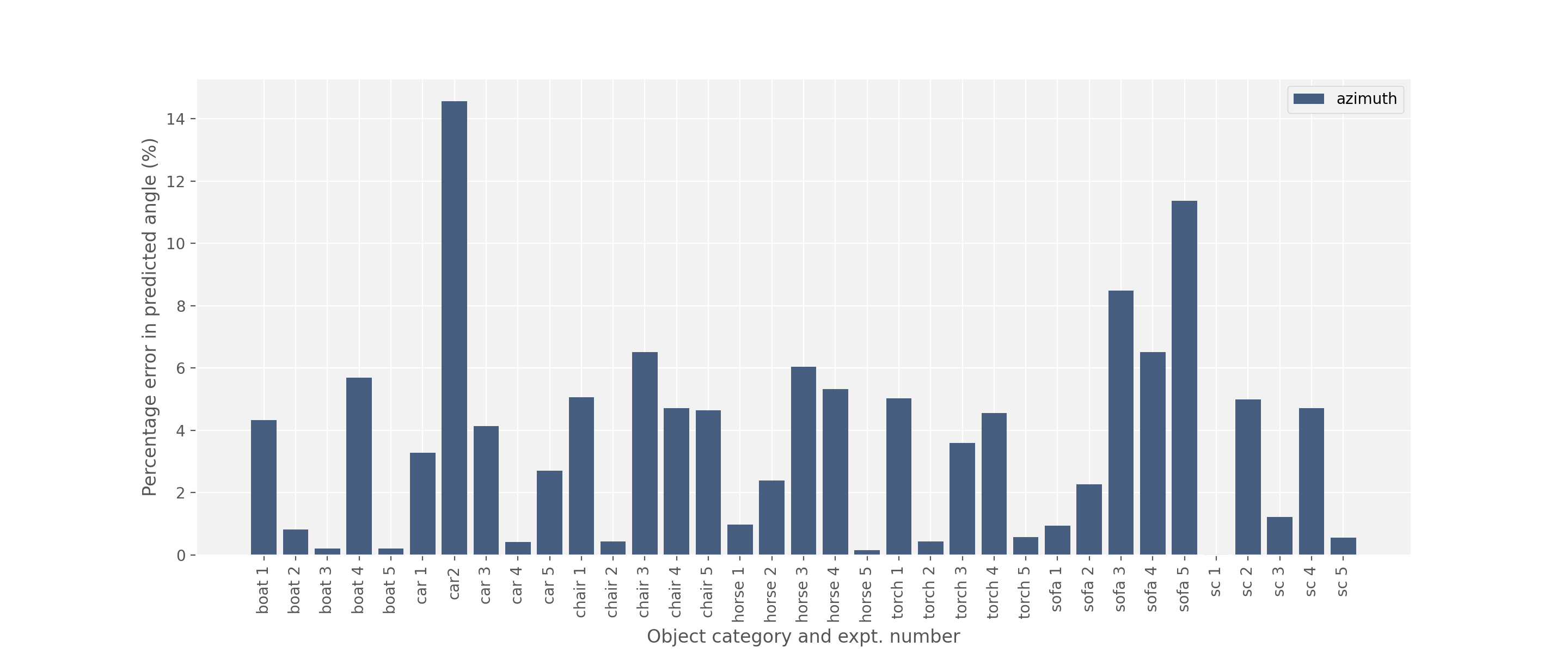}
       \caption{\textbf{Azimuth}}
        \label{fig:plot_az_percentage}
    \end{subfigure}
    \begin{subfigure}[b]{\linewidth}
       \includegraphics[width=\linewidth]{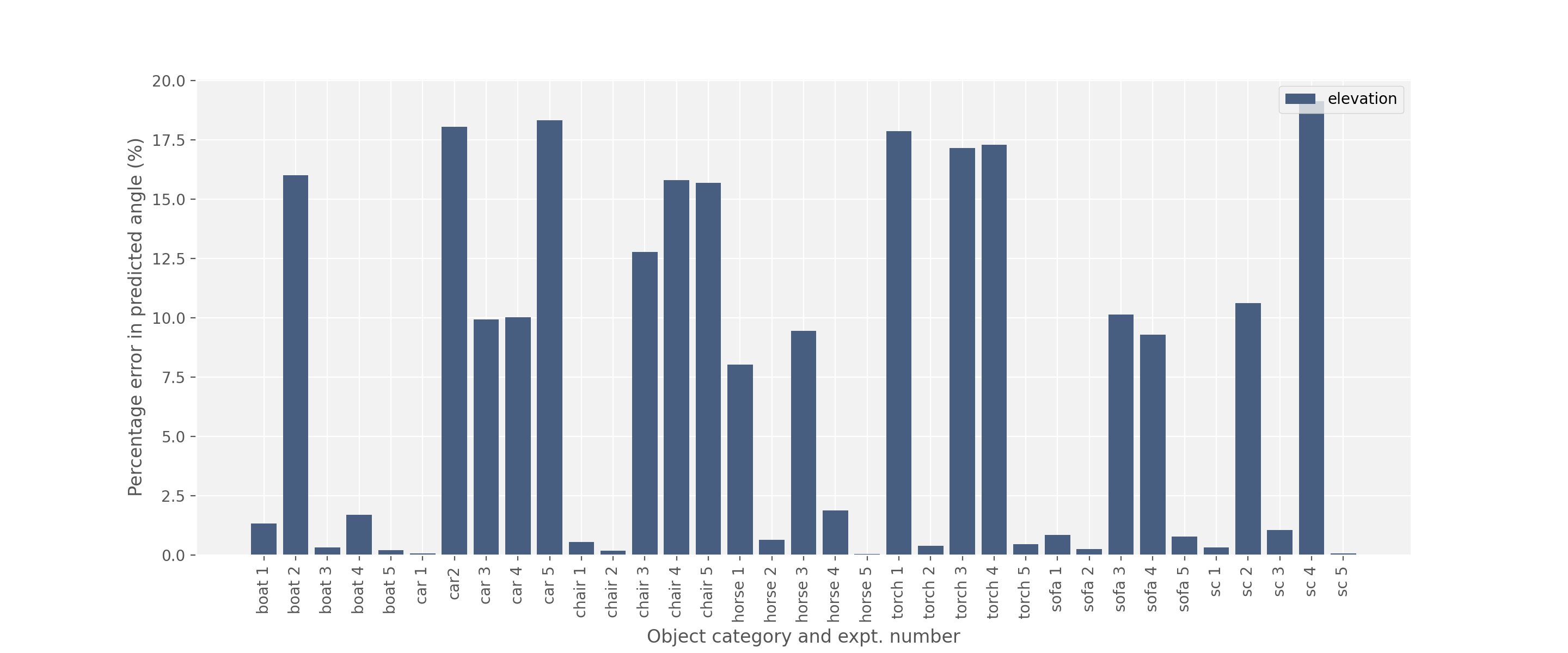}
       \caption{\textbf{Elevation}}
       \label{fig:plot_el_percentage}
    \end{subfigure}
    \begin{subfigure}[b]{\linewidth}
       \includegraphics[width=\linewidth]{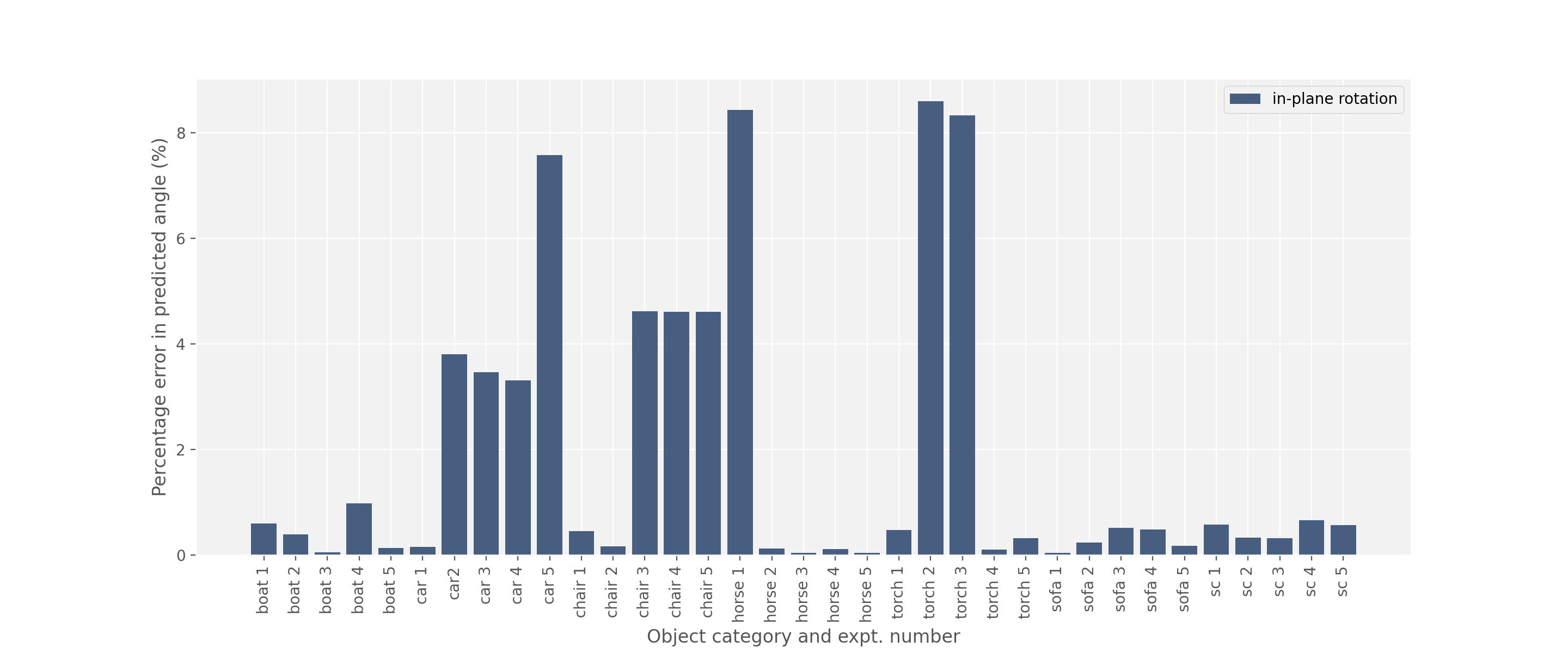}
       \caption{\textbf{In-plane rotation angle}}
       \label{fig:plot_inp_percentage}
    \end{subfigure}
    \caption{Percentage errors of predicted angles for elevation and in-plane rotation, respectively, based on best-fit azimuth predictions. Azimuth errors are calculated according to equation \ref{eq:azi_percentage_error}.}
    \label{fig:percentage_error_plots}
    \end{figure}

    Out of the total eleven 3D printed objects presented in \autoref{fig:reference_models}, pose estimation for only seven categories are shown in \autoref{fig:percentage_error_plots}. Pose estimations for the two boat models are combined in the results for the boat object category as little difference was observed for the two colors. The remaining three objects (cup, and black and white ducks) were discarded from the testing phase because they proved to be unrecognizable to the pose estimation algorithm. The cup model was discarded due to rotational symmetry, which is a common problem in pose estimation applications \cite{corona_pose_2018, hodan_epos_2020, ammirato_symgan_2020}. The pose estimation algorithm did not predict any meaningful results for the cup model (except for the elevation angle). Furthermore, we discarded the duck model due to a lack of texture. The features of both the white and the black models appeared unrecognizable to the pose estimation algorithm. In \autoref{fig:percentage_error_plots}, we present the result of several rotations along the three axis in 3D. As can be seen, the percentage error in the estimation of any angle is less than 20\%. For the in-plane rotation the estimation is particularly good with an error less than 10\%.  
    
    Our most prominent observation from the preliminary results presented in \autoref{fig:percentage_error_plots} are as follows:
    \begin{itemize}
        \item \textbf{Symmetrical objects}: 
        Detecting the pose of partly or fully symmetrical objects is a  key challenge in pose estimation. Specific measures have to be applied in order to address this challenge \cite{corona_pose_2018-1, ammirato_symgan_2020}. Our 3D cup model is not entirely symmetrical due to the handle; however, from most in-plane rotation, the object appeared to be entirely symmetrical about the rotational axis. In a digital twin context this will be the most difficult challenge to resolve. 
        \item \textbf{Shape similarities from different angles}: 
        The shapes of some objects appear similar even when seen from completely different angles. This typically poses a challenge to the pose estimation network, especially when it appears in combination with poor lighting conditions. For instance, the front and back of the 3D car model,  may appear very similar in images, making it almost impossible for the pose estimation algorithm to correctly distinguish between the different orientations. This to a certain extent explains the relatively larger error associated with the car and torch. For objects having little symmetry, the estimation was consistently more accurate. 
        \item \textbf{Objects with dark light absorbing surface}: Detecting the pose of such objects is a challenge for the 3D pose estimation network. For the pose estimation method to work, it should be able to extract relevant features. Customizing the lighting conditions to enhance such features can be one way of ensuring that the algorithm works for such objects in a digital twin. This explains the relatively higher error in estimating the elevation and in-plane rotation angles of torch and chair. 
    \end{itemize}
    
\section{Conclusion and future work}
\label{sec:conclusionandfuturework}
In the current work we present preliminary results from a workflow consisting of motion detection using DMD, object detection and classification using YOLO and pose estimation using 3D machine learning algorithm to detect rotational changes of the 3D objects in the context of digital twin. The potential of the workflow was demonstrated using a custom made experimental setup. The main conclusion from the work can be enumerated as follows:
\begin{enumerate}
    \item Although the DMD algorithm struggled to distinguish static background from the moving foregrounds clearly under challenging conditions (low light and low contrast conditions), it turned out to be a robust technique for detecting motion, a requirement for our proposed approach to geometric change detection.
    \item YOLOv5 when retrained on the images of our models had an overall mAP of 0.936. Furthermore, the algorithm proved to be very accurate in extracting the bounding box around the detected objects, a prerequisite for the pose-estimator to work in the next step.
    \item Changes in the all the three angles (azimuth, elevation and in-plane rotation) were predicted reasonably well with the error percentage well below 20\% for all the objects.  
\end{enumerate}

Having said that we admit that are several shortcomings associated with the proposed workflow. In the current work we focused only on the rotational aspect of the changes in object's orientation while ignoring the translational aspect completely. In order for the geometric change detection to be relevant for digital twin we will need information about the full 6-DOF. The problem can be solved by employing a depth camera in the setup and then using simple trigonometry to extract the translational information. Furthermore, all the trainable algorithms can be retrained on the images of objects taken under a much wider variety of lighting, texture and color condition to make the full workflow more robust. We can argue that most of the object in any digital twin will be apriori known so that full workflow can be tailored for at least those objects.
\section*{Acknowledgments}
This work was supported by the Research Council of Norway through the EXAIGON project, project number 304843


\bibliographystyle{unsrtnat}
\bibliography{references}

\end{document}